%% file: main.tex
\documentclass[10pt,twocolumn,letterpaper]{article}
\usepackage{subcaption}
\usepackage{graphicx,booktabs,makecell,array}
\usepackage{array}  % for >{} and m{}
\usepackage{tabularx} % 放在導言區
\newcolumntype{Y}{>{\centering\arraybackslash}X}
\usepackage[pagenumbers]{wacv} % To force page numbers, e.g. for an arXiv version
\usepackage{multirow}

\usepackage{microtype}


\definecolor{wacvblue}{rgb}{0.21,0.49,0.74}
\usepackage[pagebackref,breaklinks,colorlinks,allcolors=wacvblue]{hyperref}
\def\wacvPaperID{2237} % *** Enter the WACV Paper ID here
\def\confName{WACV}
\def\confYear{2026}

\title{TED-4DGS: Temporally Activated and Embedding-based Deformation for 4DGS Compression} 
\author{Cheng-Yuan Ho$^{1}$,
He-Bi Yang$^{1}$,
Jui-Chiu Chiang$^{2}$,
Yu-Lun Liu$^{1}$,
Wen-Hsiao Peng$^{1}$\\[4pt]
$^1$National Yang Ming Chiao Tung University, Taiwan\\
$^2$National Chung Cheng University, Taiwan\\[4pt]
{\tt\small\{kelvinhe0218.cs12, mrrrimge32.cs13\}@nycu.edu.tw, rachel@ccu.edu.tw}\\
\tt\small yulunliu@cs.nycu.edu.tw, wpeng@cs.nycu.edu.tw\\
}

\begin{document}

\twocolumn[{%
\renewcommand\twocolumn[1][]{#1}%
\maketitle

\includegraphics[width=1\linewidth]{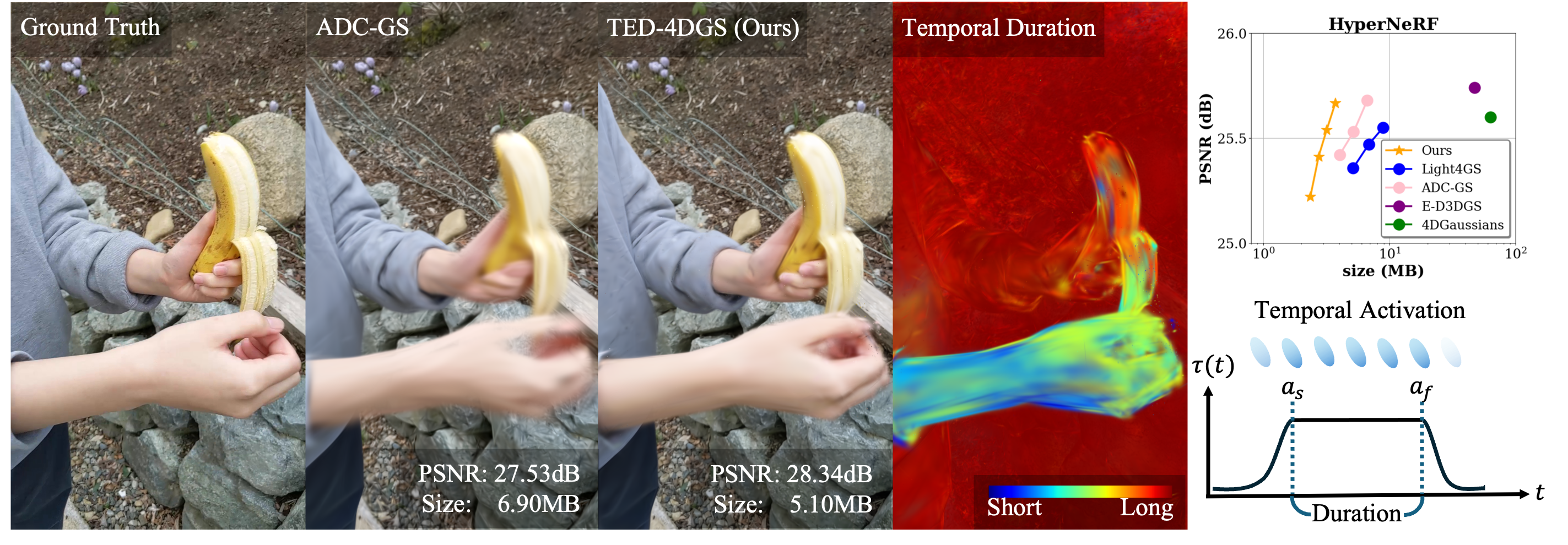}

% \vspace{-1em}
\captionof{figure}{
\textbf{Overview of TED-4DGS.} \emph{Left}: Qualitative comparison on a banana scene. Our TED-4DGS reconstructs the scene with superior rendering quality compared to ADC-GS. It achieves a 26\% file size reduction while closely matching the ground-truth view. \emph{Centre}: Temporal duration map. Static background regions reuse long-duration Gaussian primitives, whereas occluded parts of the hand and banana are represented by short-duration primitives, demonstrating the effectiveness of temporal activation. \emph{Right-top}: Rate-distortion comparison on the HyperNeRF\protect~\cite{park2021hypernerfhigherdimensionalrepresentationtopologically} benchmark. Our TED-4DGS attains higher PSNR with smaller file sizes than prior methods. \emph{Right-bottom}:Illustration of the learnable temporal-activation function, which activates a Gaussian primitive from its appearance ($a_s$) to disappearance ($a_f$).
}
\vspace{2em}
  \label{fig:teaser}
}]

\input{sec/0_abstract}    
\input{sec/intro/1_intro}
\input{sec/related_work/2_related_work}
\input{sec/3_preliminary}

\input{sec/method/4_method}

\input{sec/exp/5_exp}

\input{sec/6_conclusion}
{
    \small
    \bibliographystyle{ieeenat_fullname}
    \bibliography{main}
}

\clearpage        % <<< 把所有浮動物件排完，硬分頁

\appendix
\newpage
% \input{wacv-2026-author-kit-template/sec/Supplementary/A_Supplementary}

% {
%     \small
%     \bibliographystyle{ieeenat_fullname}
%     \bibliography{main}
% }

\end{document}

%% file: sec/0_abstract.tex
\begin{abstract}
Building on the success of 3D Gaussian Splatting (3DGS) in static 3D scene representation, its extension to dynamic scenes--commonly referred to as 4DGS or dynamic 3DGS--has attracted increasing attention. 
However, designing more compact, efficient deformation schemes together with rate-distortion-optimized compression strategies for dynamic 3DGS representations remains an underexplored area.  
Prior methods either rely on space-time 4DGS with overspecified, short-lived Gaussian primitives or on canonical 3DGS with deformation that lacks explicit temporal control. To address this, we present \textbf{TED-4DGS}, a temporally activated and embedding-based deformation scheme for rate-distortion-optimized 4DGS compression that unifies the strengths of both families. TED-4DGS is built on a sparse anchor-based 3DGS representation. Each canonical anchor is assigned with learnable temporal-activation parameters to specify its appearance and disappearance transitions over time, while a lightweight per-anchor temporal embedding queries a shared deformation bank to produce anchor-specific deformation. For rate-distortion compression, we incorporate an implicit neural representation (INR)-based hyperprior to model anchor attribute distributions, along with a channel-wise autoregressive model to capture intra-anchor correlations. With these novel elements, our scheme achieves the state-of-the-art rate-distortion performance on several commonly used real-world datasets. To the best of our knowledge, this work represents one of the first attempts to pursue a rate-distortion-optimized compression framework for dynamic 3DGS representations. 
\end{abstract}

%% file: sec/intro/1_intro.tex
\section{Introduction}
\label{sec:intro}

Reconstructing dynamic scenes from multi-view videos has long been a central challenge in 3D vision, with applications ranging from novel view synthesis and free-viewpoint rendering to dynamic scene understanding. Building on the success of 3D Gaussian Splatting (3DGS) in static 3D scene representation, its extension to dynamic scenes--commonly referred to as 4DGS or dynamic 3DGS--has attracted increasing attention. Analogous to the evolution from 2D images to 2D videos, leveraging temporal redundancy for more efficient representation calls for principled design in both temporal modeling and compression strategies. This work represents one of the first attempts to pursue a rate-distortion-optimized compression framework for dynamic 3DGS representations. 
Unlike previous methods that primarily aim to reduce memory footprint or accelerate rendering for dynamic 3DGS, our approach integrates entropy coding into the compression of Gaussian primitives for storage and transmission purposes.

%Unlike prior works dedicated to The rate-distortion optimization 

%3D Gaussian Splatting ~\cite{kerbl20233dgaussiansplattingrealtime} (3DGS) has recently emerged as a promising new representation for static 3D scenes. Subsequent works have progressively shifted their focus from reducing memory footprint and enhancing rendering speed to optimizing rate-distortion performance for efficient storage and transmission. 

%Reconstructing dynamic scenes from multi-view videos has emerged as a central problem in 3D vision, enabling applications such as novel view synthesis, free-viewpoint rendering, and dynamic scene understanding. Recently,   and existing works generally fall into two categories: the first represents the dynamic scene using explicit space-time 4DGS, while the second models deformation fields over canonical 3D Gaussian primitives to recover time-varying geometry.
%The first line of research represents dynamic scenes with space-time 4D Gaussian primitives (space-time 4DGS), 
Existing works on dynamic 3DGS representations generally fall into two categories: the first represents the dynamic scene using explicit space-time 4DGS, while the second models deformation fields over canonical 3D Gaussian primitives (i.e.~canonical 3DGS with deformation). 
Space-time 4DGS ~\cite{yang2024realtimephotorealisticdynamicscene,cho20244dscaffoldgaussiansplatting,li2024spacetimegaussianfeaturesplatting,wang2025freetimegsfreegaussianprimitives} augments the spatial attributes of 3D Gaussian primitives with temporal parameters that specify their deformation and visibility across time. 
%Methods in this category often associate each Gaussian primitive with simple parametric motion~\cite{yang2024realtimephotorealisticdynamicscene,cho20244dscaffoldgaussiansplatting,wang2025freetimegsfreegaussianprimitives,li2024spacetimegaussianfeaturesplatting} to capture its temporal evolution. They thus tend to rely on numerous short-lived Gaussian primitives to achieve high rendering fidelity. This leads to significant redundancy for compression.
In contrast, canonical 3DGS with deformation~\cite{liu2025light4gslightweightcompact4d, wu20244dgaussiansplattingrealtime, xu2024grid4d, huang2025adcgsanchordrivendeformablecompressed, bae2024pergaussianembeddingbaseddeformationdeformable} reconstructs dynamic scenes by learning deformation fields that warp canonical 3D Gaussian primitives over time. Since these methods define Gaussian primitives in a canonical space shared across time, they typically require significantly fewer Gaussian primitives compared to space-time 4DGS. Consequently, canonical 3DGS with deformation is often adopted as the core representation in rate-distortion-optimized compression frameworks~\cite{huang2025adcgsanchordrivendeformablecompressed, liu2025light4gslightweightcompact4d}. 

%Deformation signaling can be coordinate-based or embedding-based. Coordinate-based methods~\cite{xu2024grid4d,wu20244dgaussiansplattingrealtime,liu2025light4gslightweightcompact4d} use high-resolution grids to represent deformation fields and typically suffer from substantial storage requirements. In comparison, embedding-based methods associate each Gaussian primitive with a learnable embedding and decode deformation via a neural network, making the parameter count proportional to the number of Gaussian primitives. 

However, canonical 3DGS with deformation methods face a common challenge in handling occlusion and disocclusion. Since each Gaussian primitive persists throughout the entire sequence duration, the absence of temporal parameters that explicitly define its active duration may lead to peculiar deformation that is necessary, in some cases, to relocate non-contributing Gaussian primitives at specific time instances. The irregular deformation results in training instability and poses a challenge in rate-distortion compression.

To address this challenge, we propose \textbf{TED-4DGS}, a temporally activated and embedding-based deformation framework for 4DGS compression. We build TED-4DGS upon a sparse anchor-based 3DGS representation~\cite{cho20244dscaffoldgaussiansplatting}. It features a per-anchor embedding-based deformation field, where each anchor is assigned a learnable temporal feature that serves as a query to a shared global deformation bank. To model occlusion and temporal visibility, we extend canonical 3D anchors into 4D by introducing temporal activation parameters, which specify the appearance and disappearance transitions of each anchor over time. Furthermore, inspired by recent advances in 3DGS rate--distortion compression~\cite{zhan2025cat3dgscontextadaptivetriplaneapproach, chen2024hachashgridassistedcontext}, we incorporate an implicit neural representation (INR)--based hyperprior to model anchor attribute distributions, along with a channel-wise autoregressive model to capture intra-anchor correlations. In addition, we observe that cross-camera color inconsistencies across multi-view captures can severely destabilize training and degrade rendering quality. To mitigate this issue, we introduce a lightweight Color Correction Module (CCM) that compensates for camera-specific color bias during training, leading to more stable optimization and color-consistent renderings across views.

To sum up, our contributions include:
\vspace{0.5em}
\begin{itemize}
    \item We design a per-anchor embedding-based deformation network that leverages temporal features to query a shared global deformation bank, effectively capturing anchor-specific deformation.
    \vspace{0.5em}
    \item We extend static 3D anchors into 4D by introducing temporal activation parameters, promoting stable deformation and improved visibility modeling.
    \vspace{0.5em}
    \item We develop a rate-distortion-optimized compression framework that incorporates an INR-based hyperprior and a channel-wise autoregressive model for efficient attribute coding.
    \vspace{0.5em}
\end{itemize}

\noindent With these novel elements, our scheme achieves the state-of-the-art rate-distortion performance on several commonly used real-world datasets.

%% file: sec/related_work/2_related_work.tex
\section{Related Work}
\label{sec:related_work}

\input{sec/related_work/Comparison_table}

%This section reviews related work for (1) dynamic 3D Gaussian Splatting (3DGS) representations, and (2) rate-distortion-optimized 3DGS compression.

\subsection{Dynamic 3DGS Representations}

Dynamic 3DGS representations can be broadly classified into two types: (1) space-time 4DGS and (2) canonical 3DGS with deformation, as illustrated in Tab.~\ref{tab:method_compare_table}.

% 4D Gaussian based methods extend 3D Gaussian primitives

Space-time 4DGS is a natural extension of 3DGS. It augments the spatial attributes of 3D Gaussian primitives with temporal parameters that specify their motion and visibility across time. Methods in this category often feature simple parametric motion modeling, such as linear motion~\cite{yang2024realtimephotorealisticdynamicscene,cho20244dscaffoldgaussiansplatting,wang2025freetimegsfreegaussianprimitives,oh2025hybrid3d4dgaussiansplatting} or polynomial motion~\cite{li2024spacetimegaussianfeaturesplatting,lee2024fullyexplicitdynamicgaussian}. As such, recent methods in this category, e.g. 4DScaffoldGS~\cite{cho20244dscaffoldgaussiansplatting} and FreeTimeGS~\cite{wang2025freetimegsfreegaussianprimitives}, favor temporally short-lived Gaussian primitives, in order to model complex motion and enhance rendering fidelity. For instance, 4DScaffoldGS~\cite{cho20244dscaffoldgaussiansplatting} introduces a temporal coverage-aware anchor growing scheme to promote short-lived anchors, while FreeTimeGS~\cite{wang2025freetimegsfreegaussianprimitives} applies a regularization strategy to penalize Gaussian primitives with extended lifespans across time. Although showing promising rendering fidelity, they significantly increase the number of Gaussian primitives or anchors, leading to higher storage demands.
%Motion is typically modeled parametrically. Linear motion models have been adopted in 4DGS \cite{yang2024realtimephotorealisticdynamicscene}, 4DScaffoldGS \cite{cho20244dscaffoldgaussiansplatting}, and FreeTimeGS \cite{wang2025freetimegsfreegaussianprimitives}. STG \cite{li2024spacetimegaussianfeaturesplatting} adopts polynomial motion instead. 

In contrast, the notion of canonical 3DGS with deformation~\cite{kratimenos2024dynmfneuralmotionfactorization, kong2025efficientgaussiansplattingmonocular,duan20244drotorgaussiansplattingefficient,fan2025spectromotiondynamic3dreconstruction,duisterhof2024deformgssceneflowhighly,huang2024scgssparsecontrolledgaussiansplatting,labe2024dgddynamic3dgaussians,shaw2024swingsslidingwindowsdynamic,fischer2024dynamic3dgaussianfields,kwak2025modecgsglobaltolocalmotiondecomposition,yang2025ntrgaussiannighttimedynamicthermal,matsuki20254dtamnonrigidtrackingmapping,liu2025modgsdynamicgaussiansplatting,yang2024deform3dgsflexibledeformationfast} aims to reconstruct dynamic scenes by deforming a set of canonical 3D Gaussian primitives that persist throughout the entire scene duration. Deformation signaling can be either coordinate-based or embedding-based. Coordinate-based methods model the deformation field as a space-time 4D function, which is typically cast as learning a dense 4D grid. To reduce memory consumption, 4DGaussians~\cite{wu20244dgaussiansplattingrealtime} decomposes the 4D grid with a hexplane representation~\cite{cao2023hexplanefastrepresentationdynamic}, while Grid4D~\cite{xu2024grid4d} employs 3D hash grids~\cite{M_ller_2022}. Despite these efforts, coordinate-based methods still incur considerable storage overhead. 
Instead of learning grid points, E-D3DGS~\cite{bae2024pergaussianembeddingbaseddeformationdeformable}, a representative embedding-based method, learns a per-Gaussian embedding for each Gaussian primitive, along with a number of temporal embeddings, one per time instance. These embeddings are concatenated or multiplied to arrive at the deformation for a given Gaussian primitive. When coupled with a sparse canonical 3DGS representation, embedding-based methods usually have much reduced parameter count, as compared to coordinate-based methods. Although both strategies are capable of modeling complex motion, the absence of temporal parameters that explicitly define the active duration of each Gaussian primitive may lead to peculiar motion that is necessary, in some cases, to relocate non-contributing Gaussian primitives at specific time instances.

Drawing inspiration from both types of dynamic 3DGS representations, this work integrates the temporal activation capability of space-time 4DGS with the strengths of embedding-based deformation in modeling complex motion. This enables the temporal duration of each canonical 3DGS Gaussian primitive to be defined explicitly, allowing non-contributing Gaussian primitives to be excluded from rendering without resorting to unnatural motion for their relocation. Consequently, our approach achieves memory and storage efficiency by modeling complex yet natural motion with a reduced set of Gaussian primitives.

%introduce temporal parameters to embedding-based 3DGS to specify the temporal duration of each canonical 3DGS Gaussian primitive.   
%In our work, we further exploit this advantage by treating a shared temporal embedding as a motion bank, which sparse per-anchor embeddings can query to efficiently represent anchor-specific deformation. This design yields a more compact and expressive deformation representation.

%takes as input the spatial coordinates of a canonical 3D Gaussian primitive along with the timestamp at which its deformation is to be queried. Learning this 4D function
%These methods reduce parameter count, especially 

\subsection{Rate-Distortion-Based 3DGS Compression}
\label{sec:related_work_rdo}
Rate-distortion-optimized 3DGS compression is attracting growing interest due to increasing demands for efficient storage and transmission. The key challenge, which goes beyond merely reducing the parameter count of the 3DGS representation~\cite{lee2024compact3dgaussianrepresentation, fan2024lightgaussianunbounded3dgaussian, ali2024trimmingfatefficientcompression}, lies in minimizing the entropy rate of these parameters to achieve compression efficiency.

Recent work~\cite{chen2024hachashgridassistedcontext, zhan2025cat3dgscontextadaptivetriplaneapproach, wang2024contextgscompact3dgaussian,liu2025hemgshybridentropymodel,chen2025fastfeedforward3dgaussian} in 3DGS compression converges on anchor-based representations, e.g. ScaffoldGS~\cite{cho20244dscaffoldgaussiansplatting,liu2024compgsefficient3dscene}, paired with joint rate-distortion optimization via the hyperprior~\cite{ballé2018variationalimagecompressionscale} and context models~\cite{minnen2018jointautoregressivehierarchicalpriors,he2022elicefficientlearnedimage,minnen2020channelwiseautoregressiveentropymodels}. 
For rate-distortion optimization, HAC~\cite{chen2024hachashgridassistedcontext} introduces a hash grid-based hyperprior to capture spatial correlations among anchor attributes. To further exploit both inter-anchor and intra-anchor correlations, CAT-3DGS~\cite{zhan2025cat3dgscontextadaptivetriplaneapproach} employs a triplane-based hyperprior framework that features both spatial and channel-wise autoregressive modeling.

Rate-distortion-optimized compression for dynamic 3DGS scenes remains largely underexplored. Most early attempts~\cite{liu2025light4gslightweightcompact4d,huang2025adcgsanchordrivendeformablecompressed} build on canonical 3DGS representations with deformation, with a particular focus on how deformation is encoded. Light4GS ~\cite{liu2025light4gslightweightcompact4d} encodes hexplanes used for deformation modeling as images by adopting implicit neural representation (INR)-based image coding. ADC-GS~\cite{huang2025adcgsanchordrivendeformablecompressed} encodes embedding-based deformation with an autoregressive model applied along the channel dimension.

\subsection{Comparison with Prior Work}
As shown in Table~\ref{tab:method_compare_table}, our work differs from previous work on space-time 4DGS by incorporating rate-distortion optimized compression and embedding-based deformation. The capability to model complex motion via embedding-based deformation enables the representation of dynamic scenes using fewer Gaussian primitives, thereby facilitating rate-distortion optimized compression. 

In contrast to Light4GS~\cite{liu2025light4gslightweightcompact4d} and ADC-GS \cite{huang2025adcgsanchordrivendeformablecompressed}, which rely on canonical 3DGS representations with deformation, our method introduces temporal activation to more effectively manage occlusion and disocclusion without introducing artificial deformation artifacts. 

Last but not least, we replace the grid-based hyperprior, e.g.~\cite{zhan2025cat3dgscontextadaptivetriplaneapproach,chen2024hachashgridassistedcontext}, with a lightweight implicit neural representation (INR)-based hyperprior, which eliminates the signaling overhead associated with grid structures such as triplanes or hash tables.
%offers significantly improved memory efficiency and 
%\textcolor{red}{Unlike prior deformation-centric codecs, our approach adheres to the hyperprior-plus-context paradigm that has proven effective for static-scene 3DGS compression, and we treat each anchor’s temporal attributes as an additional attribute in the latent space. Specifically, we replace the grid-based hyperprior with a lightweight, pure implicit neural representation (INR) hyperprior, implemented as a position-encoded MLP that captures inter-anchor spatial correlations by directly mapping each anchor’s 3D coordinates to the mean and scale of its latent variable. This continuous formulation eliminates the fixed-grid overhead of triplanes, and is markedly more memory-efficient. Furthermore, to account for intra-anchor channel dependencies, we retain a channel-wise autoregressive entropy model, enabling fully end-to-end, joint rate–distortion optimisation for dynamic 3DGS scenes.}.

%An anchor-based approach, such as ScaffoldGS~\cite{cho20244dscaffoldgaussiansplatting,liu2024compgsefficient3dscene}, leverages a single anchor to encode a group of Gaussian primitives, dramatically reducing redundancy and enabling more compact scene modeling.

%% file: sec/related_work/Comparison_table.tex
\begin{table}[t]
\caption{Comparative analysis of dynamic 3DGS approaches.}
\label{tab:method_compare_table}
\vspace{-0.6em}
\centering
\setlength{\tabcolsep}{2pt}   % 全域欄間距稍微小一點
\renewcommand{\arraystretch}{1.0}
\footnotesize

\begin{tabular}{@{} >{\raggedright\arraybackslash}p{2.2cm}
    !{\hspace{1pt}\vrule width 0.4pt\hspace{2pt}} % 自訂直線與兩側間距
    c c c @{}}
\hline
\textbf{Model} & \textbf{Temporal Activation} & \textbf{RD Comp.} & \textbf{Motion Model}\\
\hline
\multicolumn{4}{@{}l}{\textbf{Space-time 4DGS}}\\
\hline
4DGS~\cite{yang2024realtimephotorealisticdynamicscene} & \checkmark & $\times$ & Linear\\
STG~\cite{li2024spacetimegaussianfeaturesplatting}     & \checkmark & $\times$ & Polynomial\\
4DScaffoldGS~\cite{cho20244dscaffoldgaussiansplatting} & \checkmark & $\times$ & Linear\\
FreeTimeGS~\cite{wang2025freetimegsfreegaussianprimitives} & \checkmark & $\times$ & Linear\\
\hline
\noalign{\vskip 2pt} % 往下推下一行
\multicolumn{4}{l}{\shortstack[l]{\textbf{Canonical 3DGS }\\\textbf{+ Deformation}}}\\
\noalign{\vskip 1pt} % （可選）在它與下方 \hline 之間再加一點
\hline
4DGaussians~\cite{wu20244dgaussiansplattingrealtime} & $\times$ & $\times$ & Coord.-based\\
E-D3DGS~\cite{bae2024pergaussianembeddingbaseddeformationdeformable} & $\times$ & $\times$ & Embedding-based\\
Light4GS~\cite{liu2025light4gslightweightcompact4d} & $\times$ & \checkmark & Coord.-based\\
ADC-GS~\cite{huang2025adcgsanchordrivendeformablecompressed} & $\times$ & \checkmark & Embedding-based\\
\hline
Ours (TED-4DGS) & \checkmark & \checkmark & Embedding-based\\
\hline
\end{tabular}
\end{table}

%% file: sec/3_preliminary.tex
\section{Preliminary: ScaffoldGS}

% \subsection{ScaffoldGS}
To construct an efficient 4DGS representation, our TED-4DGS leverages ScaffoldGS~\cite{lu2023scaffoldgsstructured3dgaussians}--an anchor-based 3DGS representation--as its core representation framework. ScaffoldGS ~\cite{lu2023scaffoldgsstructured3dgaussians} is a compact and storage-efficient representation initially designed for static scenes.  It introduces anchor points (referred hereafter to as anchors) located on a pre-defined voxel grid. Each anchor encapsulates information for a fixed number $K$ of Gaussian primitives. Their structural and color attributes, including the scale $\{\boldsymbol{s}_i \}_{i=0}^{K-1}$, rotation $\{\boldsymbol{r}_i \}_{i=0}^{K-1}$, color $\{\boldsymbol{c}_i \}_{i=0}^{K-1}$, and opacity $\{\alpha_i \}_{i=0}^{K-1}$, are decoded from the anchor's feature $\boldsymbol{f}$. Moreover, the anchor's position $\boldsymbol{x}$ and those $\{\boldsymbol{\mu}_i \}_{i=0}^{K-1}$ of its associated Gaussian primitives are related by $\{ \boldsymbol{\mu}_i \}_{i=0}^{K-1}= \boldsymbol{x} + \boldsymbol{l}\cdot \{ \boldsymbol{O}_i \}_{i=0}^{K-1}$, where $\{ \boldsymbol{O}_i \}_{i=0}^{K-1}$ are learned offsets, which are scaled by a learned vector $\boldsymbol{l}$. Due to its storage-friendly design, ScaffoldGS has been widely adopted in many RD-optimized compression frameworks~\cite{chen2024hachashgridassistedcontext, zhan2025cat3dgscontextadaptivetriplaneapproach, wang2024contextgscompact3dgaussian, liu2025hemgshybridentropymodel}.
%that build upon its representation to achieve high compression ratios and practical scalability.
%enabling efficient parameter sharing.
% \subsection{E-D3DGS}

% E-D3DGS \cite{bae2024pergaussianembeddingbaseddeformationdeformable}  is a deformable 3DGS framework that introduces an embedding-based approach to modeling spatio-temporal deformation. To represent spatial variation, it assigns a learnable embedding $\boldsymbol{z}_g$ to each Gaussian primitive. For temporal dynamics, it employs a shared temporal embedding $\boldsymbol{Z}_t$, which is interpolated over time.

% To obtain the deformation of a canonical Gaussian at a given timestamp $t$, E-D3DGS concatenates the per-Gaussian spatial embedding with the interpolated temporal embedding and feeds them into a deformation decoder $F_\theta$. The decoder outputs the changes in the Gaussian position, rotation, and scale.
% \[
% \Delta \boldsymbol{\mu},\ \Delta \boldsymbol{r},\ \Delta \boldsymbol{s} = F_\theta \left( \boldsymbol{z}_g,\ \text{interp}(\boldsymbol{Z}_t, t) \right).
% \]
% Unlike coordinate-based methods that model deformation fields using dense grids, E-D3DGS leverages the sparsity of Gaussian primitives through its per-Gaussian embedding design. This avoids the large storage cost associated with coordinate-based deformation fields and makes the method particularly suitable for integration with compact 3DGS representations.

%% file: sec/method/4_method.tex
\section{Proposed Method: TED-4DGS}
\input{sec/method/4_0_highlight}
\input{sec/method/4_1_System_overview}
\input{sec/method/4_2_deformation_network}
\input{sec/method/4_3_3d_to_4d}
\input{sec/method/4_4_compression}

% \input{sec/method/4_6_residual_map}
\input{sec/method/4_5_training_objectives}

%% file: sec/method/4_0_highlight.tex
Based on ScaffoldGS~\cite{lu2023scaffoldgsstructured3dgaussians}, this work introduces a temporally activated and embedding-based deformation scheme for 4DGS compression. 

First, we adopt a per-anchor embedding approach to modeling deformation.
Anchors are classified into static and dynamic types via a learnable, binary temporal mask $M_t$.
Each dynamic anchor is equipped with a temporal feature $\boldsymbol{\phi} \in \mathbb{R}^{d}$ that serves as a soft query to a global deformation bank $\boldsymbol{Z} \in \mathbb{R}^{\frac{F}{2} \times D}$ for $F$ frames, in order to retrieve its deformation. This design supports flexible motion modeling while exploiting anchor spatial sparsity to minimize the parameter count for deformation signaling.

%While prior embedding-based methods concatenate spatial and temporal embeddings, our approach uses the temporal feature as a soft query to the global motion bank to directly retrieve a deformation feature for each anchor. 
%This attention-inspired mechanism uses the query anchor to score and select only the most relevant features in the motion bank, passing this distilled context, rather than every raw feature, to the decoder and thereby achieving higher fidelity than naïve feature concatenation. 

Second, we assign each canonical anchor an explicit temporal-activation parameter $\boldsymbol{\tau}$ that defines its active duration. This facilitates more accurate modeling of occlusion and dis-occlusion, and prevents the generation of unnatural deformation caused by relocating non-contributing anchors during scene rendering at specific time instances

Lastly, we adopt an INR-based hyperprior framework to model the coding distributions of anchor attributes, including the anchor's feature $\boldsymbol{f}$, scaling $\boldsymbol{l}$, offsets $\{\boldsymbol{O}_i\}_{i=0}^{K-1}$, temporal feature $\boldsymbol{\phi}$, and temporal activation $\boldsymbol{\tau}$. 

%For coding $\boldsymbol{f}$, which carries much structural and color information,  we further incorporate a channel-wise autoregressive model to exploit intra correlations among its components. These attributes are jointly compressed in an rate-distortion-optimized manner.

%presents a temporal activated 4D Gaussian splatting framework that integrates an embedding-based deformation network with a learned compression pipeline.

%% file: sec/method/4_1_System_overview.tex
\subsection{System Overview}
\input{sec/method/system_overview}
Fig.~\ref{fig:overview} illustrates our TED-4DGS framework. The representation of a dynamic scene begins with the generation of anchor points, each characterized by a position and a set of spatial and temporal attributes. The spatial attributes, adopted from ScaffoldGS ~\cite{lu2023scaffoldgsstructured3dgaussians}, capture the geometry and appearance of Gaussian primitives in canonical space, while the temporal attributes encode deformation (; Section~\ref{sec:method/deformation}). To generate an anchor-specific deformation latent $\boldsymbol{z}_{t}^{a}$ for time instance $t$, the temporal feature $\boldsymbol{\phi}$ of a dynamic anchor is multiplied by a global deformation vector $z^{(t)}$, which is obtained via temporal interpolation between deformation vectors corresponding to uniformly sampled time instances in the global deformation bank $\boldsymbol{Z}$. The resulting latent $\boldsymbol{z}_{t}^{a}$ is fed into the deformation decoder $F_{\text{deform}}$ to predict 
an anchor displacement $\Delta\boldsymbol{x}$ and a feature residual $\Delta\boldsymbol{f}$. These are used to update $\boldsymbol{x}$ and $\boldsymbol{f}$ as the deformed anchor position $\boldsymbol{x}'$ and feature $\boldsymbol{f}'$, respectively. Finally, $\boldsymbol{f}'$ is used to decode the scale 
$\{\boldsymbol{s}_i\}_{i=0}^{K-1}$ and rotation $\{\boldsymbol{r}_i\}_{i=0}^{K-1}$ for each Gaussian primitive, 
whereas the color $\{\boldsymbol{c}_i\}_{i=0}^{K-1}$ and opacity $\{\alpha_i\}_{i=0}^{K-1}$ 
are treated as time-invariant and are decoded from the canonical feature $\boldsymbol{f}$.
%$\boldsymbol{x}'=\boldsymbol{x}+\Delta\boldsymbol{x}$ and 
%$\boldsymbol{f}'=\boldsymbol{f}+\Delta\boldsymbol{f}$, respectively. 
%The temporal mask $M_t$ indicates whether an anchor is static or dynamic, and the temporal activation $\boldsymbol{\tau}$ (Section ~\ref{sec:method/3d_to_4d}) defines the active duration of an anchor. These temporal attributes enable anchor-wise deformation across time.

The rendering process utilizes both dynamic and static anchors. Dynamic anchors are temporally deformed by decoding their respective temporal attributes. With the deformed dynamic anchors and static anchors, the remaining procedure largely follows ScaffoldGS, except that each Gaussian primitive’s opacity is additionally modulated by the temporal activation of its associated anchor to account for temporal visibility. 

%, such as $\boldsymbol{\phi}$, $\boldsymbol{M}_t$ and $\boldsymbol{\tau}$, global deformation bank $Z$ and the deformation decoder $F_{deform}$.

%T, and follows mostly ScaffoldGS.
%After decoding the deformed anchors,  The only difference is that we multiply 

The entropy encoding (Section~\ref{sec:method/compression}) of spatial and temporal features for each anchor leverages an INR-based hyperprior. It applies positional encoding to the anchor's position, with the resulting representation used to derive quantization step sizes and Gaussian parameters (i.e. means and variances) for entropy encoding associated attributes. Notably, for coding the anchor feature $\boldsymbol{f}$, which carries much structural and color information,  we further incorporate a channel-wise autoregressive model to exploit intra correlations among its components. These attributes are jointly compressed in an rate-distortion-optimized manner. 
Additionally, an offset mask $\boldsymbol{M}_o$ is learned to suppress less critical Gaussian primitives.
%It takes as input an anchor's position and outputs the quantization parameter
%is employed to model the distribution of anchor attributes. 
%Given an anchor position, we apply positional encoding and use it to decode the distribution of . Additionally, a channel-wise autoregressive model is applied to the anchor feature to improve compression efficiency further. 
%we define anchor attributes as the combination of spatial attributes (anchor feature $\boldsymbol{f}$, scaling $\boldsymbol{l}$, offsets $\{{\boldsymbol{O}}_i\}$) and temporal attributes (temporal feature $\boldsymbol{\phi}$ and activation $\boldsymbol{\tau}$). These attributes are all compressed in a rate-distortion-optimized manner.

The final compressed bitstream include (a) anchors' positions $\boldsymbol{x}$ and attributes, (b) the global deformation bank, (c) the network weights of the deformation decoder, Scaffold MLP decoder, hyperprior decoder, and channel-wise autoregressive model, and (d) the binary offset and temporal masks. The anchors' positions are stored in 16-bit floating-point format (FP16), while the network weights and the global deformation bank are in FP32. Both masks are entropy encoded.

%% file: sec/method/system_overview.tex
% \begin{figure*}
%   \centering
%   \begin{subfigure}{0.9\linewidth}
%     \includegraphics[width=\linewidth]{Figures/Overview_teaser/Overview.png}
%     % \caption{Deformation framework.}
%     % \label{fig:overview_deformation}
%   \end{subfigure}
%   % \hfill
%   % \begin{subfigure}{0.22\linewidth}
%   %   \includegraphics[width=\linewidth]{Figures/Overview_teaser/Compress_Overall.png}
%   %   \caption{Compression framework.}
%   %   \label{fig:overview_compression}
%   % \end{subfigure}
%   \vspace{-1em}
%   \caption{System overview of our TED-4DGS framework. The \emph{Left Top} shows the overall pipeline, while the \emph{Right Top} and \emph{Down} parts provide detailed views of the compression and deformation frameworks, respectively.
% }
%   \label{fig:overview}
% \end{figure*}

\begin{figure*}[t]
\vspace{-0.25em}
  \centering
  % 第一張圖
  \begin{subfigure}{0.87\linewidth}
    \centering
    \includegraphics[width=\linewidth]{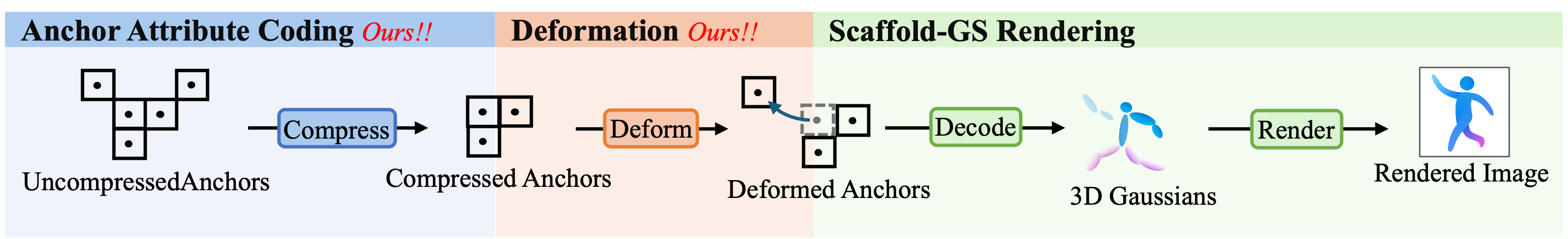}
    \vspace{-1.25em}
    \caption{Overall pipeline.}
    \label{fig:overview_pipeline}
  \end{subfigure}
  % 第二張圖
  \begin{subfigure}{0.87\linewidth}
    \centering
    \includegraphics[width=\linewidth]{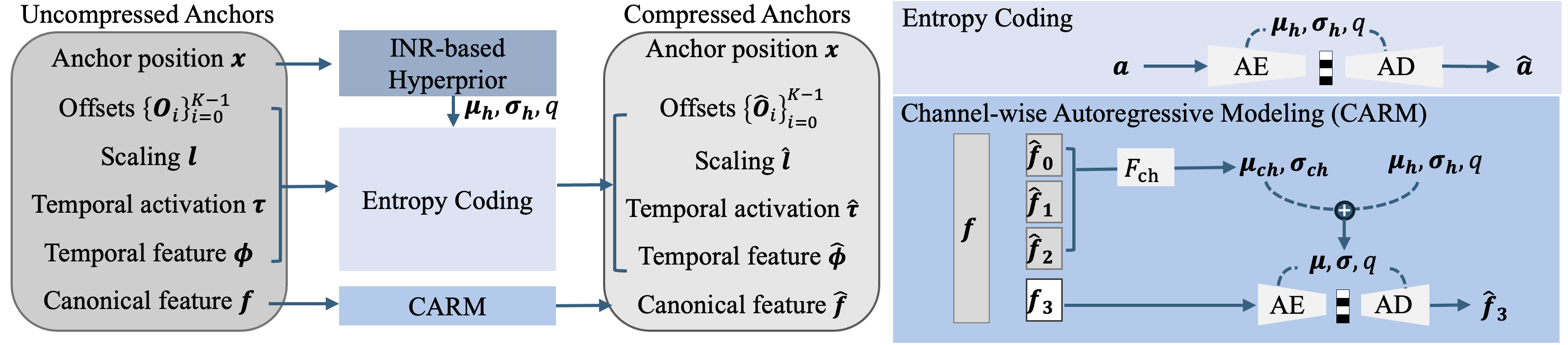}
    \vspace{-1.25em}
    \caption{INR-based anchor attribute compression framework.}
    \label{fig:overview_compression}
  \end{subfigure}
\vspace{0.25em}
  \begin{subfigure}{0.87\linewidth}
    \centering
    \includegraphics[width=\linewidth]{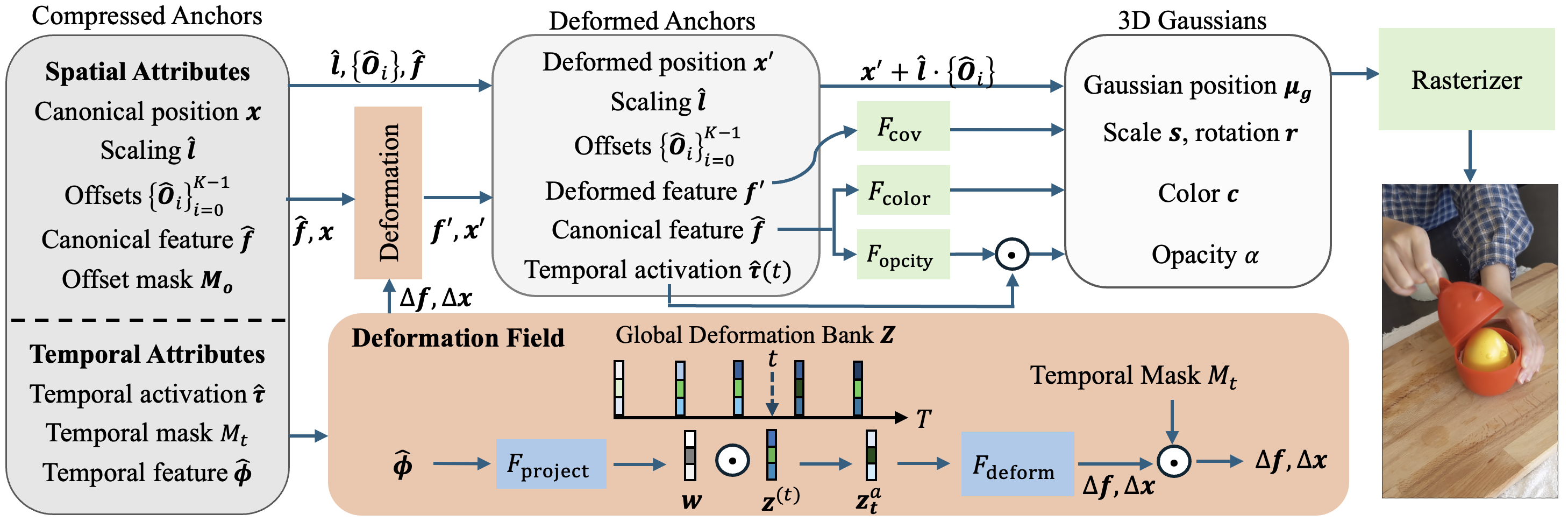}
    \vspace{-1.25em}
    \caption{Anchor-based deformation framework and Scaffold-GS rendering.}
    \label{fig:overview_deformation}
  \end{subfigure}
\vspace{-0.5em}
  \caption{System overview of our TED-4DGS framework.}
  \vspace{-1em}
  \label{fig:overview}
\end{figure*}

%% file: sec/method/4_2_deformation_network.tex
\subsection{Deformation Fields}
\label{sec:method/deformation}
To model dynamic motion, we introduce an embedding-based deformation field that predicts the temporal deformation of each anchor (Fig.~\ref{fig:overview_deformation}. The deformation is conditioned on both the per-anchor temporal feature $\boldsymbol{\phi} \in \mathbb{R}^{d}$ and a global deformation bank $\boldsymbol{Z} \in \mathbb{R}^{\frac{F}{2} \times D}$ for $F$ frames. In video language, $\boldsymbol{Z}$ functions analogously to a set of optical flow maps while $\boldsymbol{\phi}$ serves as a pixel's location that determines where the relevant motion of that pixel should be extracted from these flow maps.

The deformation inference for a dynamic anchor at time $t$ starts by interpolating temporally between deformation vectors for various time instances in the global 
deformation bank $\boldsymbol{Z}$. The outcome $\boldsymbol{z}^{(t)}=\text{interp}(\boldsymbol{Z}, t)$ is a time-specific $\boldsymbol{z}^{(t)} \in \mathbb{R}^{D}$, which (similar to a temporally-interpolated optical flow map in the context of video)
encapsulates deformation information for all dynamic anchors at time $t$ and serves as a shared query target. To perform the query, the temporal feature $\boldsymbol{\phi}$ of a dynamic anchor is projected by $\boldsymbol{w}=F_{\text{project}}( \boldsymbol{\phi})$, which is multiplied with $\boldsymbol{z}^{(t)}$ to yield an anchor-specific temporal latent $\boldsymbol{z}_{t}^{a} = \boldsymbol{w} \cdot \boldsymbol{z}^{(t)}$. This latent $\boldsymbol{z}_{t}^{a}$ further undergoes a deformation decoder $F_{\text{deform}}$ to arrive at an positional displacement $\Delta \boldsymbol{x}$ and a feature residual $\Delta \boldsymbol{f}$:
\begin{equation}
  \Delta\boldsymbol{f},\ \Delta\boldsymbol{x}
  = F_{\text{deform}}\bigl(
      F_{\text{project}}(\boldsymbol{\phi})\;
      \cdot\;
      \operatorname{interp}(\boldsymbol{Z}, t)
    \bigr),
  \label{eq:deform}
\end{equation}
We then update the anchor’s position and feature as
$\boldsymbol{x}' = \boldsymbol{x} + \Delta\boldsymbol{x}$ and $\boldsymbol{f}' = \boldsymbol{f} + \Delta\boldsymbol{f}$.

We note that the deformation vectors stored in $\boldsymbol{Z}$ are learnable parameters that must be explicitly signaled in the bitstream. To balance deformation accuracy and bitrate, we signal a deformation vector for every other frame. 
%where \( F_{\text{project}}( \boldsymbol{\phi}) \) projects the temporal feature $\boldsymbol{\phi}$, \( \text{interp}(\boldsymbol{Z}, t) \) provides the interpolated global latent at time \( t \), and \( F_{\text{deform}}  \) is the deformation decoder. 
%The deformed feature \( \boldsymbol{f}' = \boldsymbol{f} + \Delta \boldsymbol{f} \) is used to decode the covariance (i.e. scale $\{\boldsymbol{s}_i\}_{i=0}^{K-1}$ and rotation $\{\boldsymbol{r}_i\}_{i=0}^{K-1}$) of the corresponding Gaussian primitive, while the color and opacity remain temporally invariant and are decoded directly from the canonical anchor feature.

% Since not all anchors undergo temporal motion, we introduce a binary temporal mask that determines whether an anchor is dynamic or static. If an anchor is marked as static, it retains its canonical position and feature without undergoing deformation.

%% file: sec/method/4_3_3d_to_4d.tex
\subsection{Progressive 3D-to-4D Anchor Transformation}
\label{sec:method/3d_to_4d}

A key challenge in deformation-based dynamic scene representations lies in handling occlusion and disocclusion, as canonical anchors persist throughout the entire sequence duration. Ideally, when an object is occluded by another, its corresponding anchors should remain behind those of the occluder. However, as shown in Fig.~\ref{fig:3d_4d_pc}, a trivial yet improper solution is to deform occluded anchors such that they fall outside the viewing frustum. This leads to a highly irregular deformation field, hindering compression efficiency.
%However, since the deformation field is only supervised by image-based rendering losses.  This often results in unstable deformation behavior and noisy temporal dynamics.

To address this issue, we extend the static 3D anchor representation into a 4D formulation by introducing a learnable temporal activation parameter $\tau = [(a_s, b_s), (a_f, b_f)]$, which explicitly specifies each anchor's active duration as
\begin{equation}
\tau(t)=
\begin{cases}
\exp\!\left[-\bigl((t-a_s)/b_s\bigr)^2\right], & t < a_s,\\[4pt]
1, & a_s \le t \le a_f,\\[4pt]
\exp\!\left[-\bigl((t-a_f)/b_f\bigr)^2\right], & t > a_f,
\end{cases}
\label{eq:temporal_activation}
\end{equation}
where \( (a_s, b_s) \) and \( (a_f, b_f) \) define the smooth transitions for appearance and disappearance, respectively.
The specification of $\tau(t)$ allows each anchor and its associated Gaussian primitives to activate or deactivate gradually over time. During rendering, the time-aware opacity \( \alpha_t \) for each Gaussian primitive at time \( t \) is computed by modulating its static opacity \( \alpha \) with the temporal activation \( \tau(t) \):
% \vspace{-0.25em}
\begin{equation}
  \alpha_t = \alpha \cdot \,\tau(t),
  \label{eq:opacity_at_t}
\end{equation}
% \vspace{-0.75em}
which enables dynamic visibility modeling over time.

To ensure stable learning, we adopt a progressive training strategy. During the initial 20{,}000 iterations, all anchors are treated as static 3D anchors without temporal activation modeling, encouraging them to remain consistently active across frames. Subsequently, the temporal activation parameters $a_s,a_f$ are initialized for each anchor based on its earliest and latest time instances that an anchor is visible in the viewing frustum. These temporal parameters are then optimized jointly with other anchor attributes.

%\paragraph{Temporal-aware opacity based anchor pruning.} 
As a side note, we adopt the anchor pruning strategy of ScaffoldGS~\cite{lu2023scaffoldgsstructured3dgaussians}, removing anchors whose associated Gaussian primitives exhibit negligible collective opacity. By incorporating temporal activation, our time-aware opacity enables more effective removal of anchors that contribute minimally over time.

%In ScaffoldGS~\cite{lu2023scaffoldgsstructured3dgaussians}, anchor pruning is performed by accumulating the opacity of its associated Gaussian primitives; anchors are removed if their primitives do not contribute sufficient opacity to the rendered scene. In the 3D anchor setting, opacity remains fixed over time, making it difficult to identify anchors that appear dense but lie outside the viewing frustum for most of the sequence. 

%% file: sec/method/4_4_compression.tex
\subsection{INR-based Compression for Anchor Attributes}
\label{sec:method/compression}
To encode anchors' attributes in a rate-distortion-optimized fashion, we propose an INR-based hyperprior to predict the distribution parameters for entropy coding and a channel-wise autoregressive model to further exploit intra-anchor correlations for its feature compression.

%\paragraph{INR-based Hyperprior.}
The INR-based hyperprior is implemented as a multilayer perceptron that learns prior distributions over anchor attributes, including the feature $\boldsymbol{f}$, offsets $\{\boldsymbol{O}_i\}_{i=0}^{K-1}$, scaling $\boldsymbol{l}$, temporal feature $\boldsymbol{\phi}$ and temporal activation $\boldsymbol{\tau}$. Given the positional embedding of the anchor’s position $\boldsymbol{x}$, the network outputs the corresponding distribution parameters: means $ \boldsymbol{\mu}_h$, variances $\boldsymbol{\sigma}_h$, and quantization step sizes $q$. For each attribute type, a distinct quantization step size is predicted and shared across its components. The probability of a quantized attribute \( \hat{\boldsymbol{a}} \) is evaluated as
\begin{equation}
  p(\hat{\boldsymbol{a}} \mid \boldsymbol{x}\bigr)
  = \int_{\hat{\boldsymbol{a}} - \frac{q}{2}}^{\hat{\boldsymbol{a}} + \frac{q}{2}}
      \mathcal{N}\!\bigl(\boldsymbol{\mu}_h,\;\boldsymbol{\sigma}_h\bigr)\,d\boldsymbol{a},
  \label{eq:prob_hyper}
\end{equation}
which amounts to the likelihood of the quantized value under the learned Gaussian prior. Since an INR hyperprior is grid-free, we transmit only a few network weights, eliminating the extra bits that triplanes or hash tables require and keeping the model lightweight.
%\paragraph{Channel-wise Autoregressive Model.}

Following~\cite{zhan2025cat3dgscontextadaptivetriplaneapproach}, we adopt a channel-wise autoregressive model to exploit intra-feature dependencies for coding anchor features $\boldsymbol{f}$, which typically comprise a substantial portion of the bitstream (Fig.~\ref{fig:overview_compression}). 

%Specifically, \( \boldsymbol{f} \) is divided into \( M \) channel-wise slices, and an autoregressive model $F_{\text{ch}}$ is used to model the distribution of each slice based on previously encoded ones. For slice \( \boldsymbol{f}_m \) (with \( m > 0 \)), the conditional probability is defined as:
%\begin{equation}
%\begin{aligned}
%  p(\hat{\boldsymbol{f}}_m \mid \boldsymbol{x},\,\hat{\boldsymbol{f}}_{0:m-1}\bigr)
%  &= \int_{\hat{\boldsymbol{f}}_m - \frac{q}{2}}^{\hat{\boldsymbol{f}}_m + \frac{q}{2}}
     %\mathcal{N}\!\bigl(\boldsymbol{\mu}, \boldsymbol{\sigma}\bigr)\,d\boldsymbol{f}, \\[6pt]
  %\text{where}\quad
  %\boldsymbol{\mu} &= %\boldsymbol{\mu}_h + %\boldsymbol{\mu}_{ch},\quad
  %\boldsymbol{\sigma} = \boldsymbol{\sigma}_h + \boldsymbol{\sigma}_{ch}.
%\end{aligned}
%\label{eq:prob_carm}
%\end{equation}

%\paragraph{Offset pruning.}

%Based on ~\cite{chen2024hachashgridassistedcontext, lee2024compact3dgaussianrepresentation}, we introduce a binary offset mask that prunes offsets with negligible contributions. For each offset, the mask is predicted through a learnable parameter and binarized using a straight-through estimator, allowing gradient-based optimization during training. 
% To guide the learning of this mask, we additionally incorporate a mask loss that encourages sparsity, promoting more compact anchor representations and improving overall compression efficiency.

%% file: sec/method/4_5_training_objectives.tex
\subsection{Training Objectives}
\label{sec:training_objectives}

The training objective of TED-4DGS, as defined in Eq.~\eqref{eq:loss_func}, comprises several terms with distinct roles. 
The distortion loss $\mathcal{L}_{\text{distortion}}$ and rate loss $\mathcal{L}_{\text{rate}}$
form the classical rate-distortion pair, where $\mathcal{L}_{\text{distortion}}$ combines L1 and SSIM losses and $\mathcal{L}_{\text{rate}}$ denotes the average bitrate per anchor.
To promote sparsity, we adopt the offset mask loss $\mathcal{L}_{\text{offset mask}}$ and temporal mask loss $\mathcal{L}_{\text{temp mask}}$ from~\cite{lee2024compact3dgaussianrepresentation}. These losses suppress redundant spatial offsets and identify static anchors, respectively. The weight of $\mathcal{L}_{\text{offset mask}}$ is scaled proportionally to the rate-control factor $\lambda_{\text{rate}}$, so that lowering the target bitrate encourages more aggressive pruning of anchors and reduce the model size. 
Lastly, the regulization losses $\mathcal{L}_{\text{vol}}$ and $\mathcal{L}_{\text{tv}}$
encourage scale consistency and enforce temporal smoothness on the deformation vectors in the deformation bank $\boldsymbol{Z}$. These are adopted from ScaffoldGS \cite{lu2023scaffoldgsstructured3dgaussians} and E-D3DGS \cite{bae2024pergaussianembeddingbaseddeformationdeformable}. The complete objective is thus:
\begin{align}
\mathcal{L}  &=  \mathcal{L}_{\text{distortion}}
             + \lambda_{\text{rate}}
               \bigl( \mathcal{L}_{\text{rate}}
                   + \lambda_{\text{offset mask}}\,
                     \mathcal{L}_{\text{offset mask}}\bigr) \notag\\
            &\quad
             + \lambda_{\text{temp mask}}\, \mathcal{L}_{\text{temp mask}}
             + \lambda_{\text{vol}}\,  \mathcal{L}_{\text{vol}}
             + \lambda_{\text{tv}}\, \mathcal{L}_{\text{tv}}.
\label{eq:loss_func}
\end{align}
%where $\mathcal{L}_{\text{distortion}}$ combines L1 and SSIM losses. $\mathcal{L}_{\text{rate}}$ denotes the average bitrate per anchor, estimated by the entropy models in our compression framework. $\mathcal{L}_{\text{offset mask}}$ and $\mathcal{L}_{\text{temp mask}}$ are the masking losses adopted from~\cite{lee2024compact3dgaussianrepresentation}. They are used to promote sparsity in spatial offsets and identify static anchors, respectively. We scale the weight of $\mathcal{L}_{\text{offset mask}}$ proportionally to the rate-control factor $\lambda_{\text{rate}}$, so that lowering target bitrates encourages more aggressive pruning of anchor points and, in turn, reduce the overall model size. Following ScaffoldGS \cite{lu2023scaffoldgsstructured3dgaussians} and E-D3DGS \cite{bae2024pergaussianembeddingbaseddeformationdeformable}, $\mathcal{L}_{\text{vol}}$ regularizes Gaussian sizes, while $\mathcal{L}_{\text{tv}}$ enforces temporal smoothness on the global deformation bank $\boldsymbol{Z}$.

%% file: sec/exp/5_exp.tex
\section{Experiments}
\input{sec/exp/5_0_experiments}
\input{sec/exp/5_1_ablation}

%% file: sec/exp/5_0_experiments.tex
\input{sec/exp/rd_curve.tex}
\label{sec:exp/experiments}
 \paragraph*{\textbf{Implementation Details.}} Our method is implemented in PyTorch and trained on an NVIDIA RTX 3090 GPU. We evaluate a range of rate-control factors $\lambda_{\text{rate}}$, choosing $\lambda_{\text{rate}}=\{0.001, 0.002, 0.004, 0.008\}$ for the HyperNeRF dataset and $\lambda_{\text{rate}}=\{0.002, 0.004, 0.008, 0.016\}$ for the Neu3D dataset. In our experiments, the low-rate configuration corresponds to the highest $\lambda_{\text{rate}}$, which yields the lowest bitrate, whereas the high-rate configuration uses the lowest $\lambda_{\text{rate}}$ to achieve the maximum bitrate. The details of training process and the choice of hyperparameters are provided in the supplementary document.
 %The training process is divided into two stages. In the first stage, we jointly optimize the canonical representation and deformation field over 60{,}000 iterations, with anchor densification enabled. A progressive learning strategy is employed: during the first 20{,}000 iterations, anchors are trained as 3D representations; from iteration 20{,}000 onward, temporal activation modeling is introduced to enable 4D anchor learning. Importantly, rate-related objectives such as \( \mathcal{L}_{\text{rate}} \) and \( \mathcal{L}_{\text{offset mask}} \) are not included in this stage. In the second stage, we perform joint rate-distortion optimization by training the hyperprior and autoregressive models while enabling the complete training objective. We evaluate multiple $\lambda_{\text{rate}}$ values. We choose $\lambda_{\text{rate}}=\{0.001, 0.002, 0.004, 0.008\}$ for the HyperNeRF dataset and $\lambda_{\text{rate}}=\{0.002, 0.004, 0.008, 0.016\}$ for the Neu3D dataset. More implementation details are provided in {\color{red} Appendix}.

\vspace{-1em}
\paragraph*{\textbf{Baselines.}}
We compare TED-4DGS with three groups of baseline methods: (1) rate-distortion-optimized compression approaches, including Light4GS~\cite{liu2025light4gslightweightcompact4d} and ADC-GS~\cite{huang2025adcgsanchordrivendeformablecompressed}, and non-rate-distortion-optimized approaches, including (2) deformation-based methods--e.g.  4DGaussians~\cite{wu20244dgaussiansplattingrealtime} and E-D3DGS~\cite{bae2024pergaussianembeddingbaseddeformationdeformable}, and (3) space-time 4DGS methods--e.g. 4DGS~\cite{yang2024realtimephotorealisticdynamicscene}, STG~\cite{li2024spacetimegaussianfeaturesplatting} and FreeTimeGS~\cite{wang2025freetimegsfreegaussianprimitives}. 
% Note that the 4D Gaussian-based methods are only evaluated on the multi-view dataset, as they are not designed to handle monocular or stereo scenarios.

\vspace{-1em}
\paragraph*{\textbf{Datasets.}}
We follow the common test protocol~\cite{huang2025adcgsanchordrivendeformablecompressed} to evaluate our TED-4DGS on two real-world datasets: Neural 3D Video (Neu3D)~\cite{li2022neural3dvideosynthesis} and HyperNeRF~\cite{park2021hypernerfhigherdimensionalrepresentationtopologically}. Specifically, on the Neu3D dataset, which includes multiple synchronized multi-view videos captured by 18–21 cameras per scene, the results are reported for the \textit{cook spinach}, \textit{cut roasted beef}, \textit{flame salmon}, \textit{flame steak}, and \textit{sear steak} sequences. On the HyperNeRF dataset, which includes videos captured using two phones rigidly mounted on a handheld stereo rig.~\cite{park2021hypernerfhigherdimensionalrepresentationtopologically}, we report results for four dynamic scenes, \textit{3D printer}, \textit{banana}, \textit{broom}, and \textit{chicken}). We train (and test) on all frames at half resolution (536×960) in width and height according to ~\cite{huang2025adcgsanchordrivendeformablecompressed,wu20244dgaussiansplattingrealtime,yang2023deformable3dgaussianshighfidelity}. 

%At test time, and further downsample inputs by a factor of four following the test protocol in~\cite{huang2025adcgsanchordrivendeformablecompressed}.
\vspace{-1em}
\paragraph*{\textbf{Metrics.}}
To compare the rate-distortion (RD) performance of the competing methods, we report peak-signal-to-noise ratio (PSNR), structural similarity index (SSIM$_2$), and perceptual quality measure LPIPS. For the rate-distortion-optimized approaches, we compute the average PSNR/SSIM$_2$/LPIPS and compressed file size across sequences for the low-rate and high-rate configurations. Likewise, we measure rendering speed in frames per second (FPS) under these configurations.

\subsection{Rate-Distortion Comparison}

\input{sec/exp/quantitive_results}
\input{sec/exp/quality_results_2scene}

%\paragraph*{\textbf{Quantitative Results.}}
In Fig.~\ref{fig:rd_curve}, Table~\ref{tab:neu3d-results} and Table~\ref{tab:hypernerf-results}, our TED-4DGS consistently outperforms the competing methods across both datasets, achieving the state-of-the-art RD performance. 

Figs.~\ref{fig:render} further presents the subjective quality comparison. On the \textit{sear steak} scene (Neu3D), our TED-4DGS delivers similar or superior rendering quality while reducing the file size by over 14x relative to E-D3DGS~\cite{bae2024pergaussianembeddingbaseddeformationdeformable} and over 18x compared to 4DGaussians~\cite{wu20244dgaussiansplattingrealtime}.
% In Fig.~\ref{fig:rd_curve}, Table~\ref{tab:neu3d-results} and Table~\ref{tab:hypernerf-results}, our TED-4DGS consistently outperforms the competing methods across both datasets, achieving the state-of-the-art RD performance. 

% Figs.~\ref{fig:render} further presents the subjective quality comparison\footnote{Light4GS, an rate-distortion-optimized approach, is excluded from the subjective comparison due to the unavailability of its code. Likewise, the reconstructed results for ADC-GS are omitted on the Neu3D dataset, as its code could not be executed successfully on that dataset.}. On the \textit{sear steak} scene (Neu3D), our TED-4DGS delivers similar or superior rendering quality while reducing the file size by over 14x relative to E-D3DGS~\cite{bae2024pergaussianembeddingbaseddeformationdeformable} and over 18x compared to 4DGaussians~\cite{wu20244dgaussiansplattingrealtime}.

On the \textit{3D printer} scene (HyperNeRF), our TED-4DGS achieves 23.1dB PSNR with a file size of 3.4MB, which amounts to a 28\% bitrate reduction compared to the rate-distortion-optimized ADC-GS~\cite{huang2025adcgsanchordrivendeformablecompressed} at a comparable perceptual quality level. Compared to E-D3DGS~\cite{bae2024pergaussianembeddingbaseddeformationdeformable}, a non-rate-distortion-optimized approach, the reduction in file size exceeds 14x.

%among ~\cite{bae2024pergaussianembeddingbaseddeformationdeformable,wu20244dgaussiansplattingrealtime, huang2025adcgsanchordrivendeformablecompressed} 

%% file: sec/exp/rd_curve.tex
% \begin{figure}[t]
%     \centering
%     \resizebox{\linewidth}{!}{
%     \includegraphics[0.96\columnwidth]{Figures/exp/rd.png}  
%     }
%     \vspace{-2em}
%     \caption{Rate-distortion comparison of our TED-4DGS, Light4GS~\cite{liu2025light4gslightweightcompact4d} and ADC-GS~\cite{huang2025adcgsanchordrivendeformablecompressed}.}
%     \label{fig:rd_curve}
%     \vspace{-1em}
% \end{figure}

\begin{figure}[t]
  \centering
  \begin{subfigure}[t]{0.48\columnwidth}
      \centering
      \includegraphics[width=\linewidth]{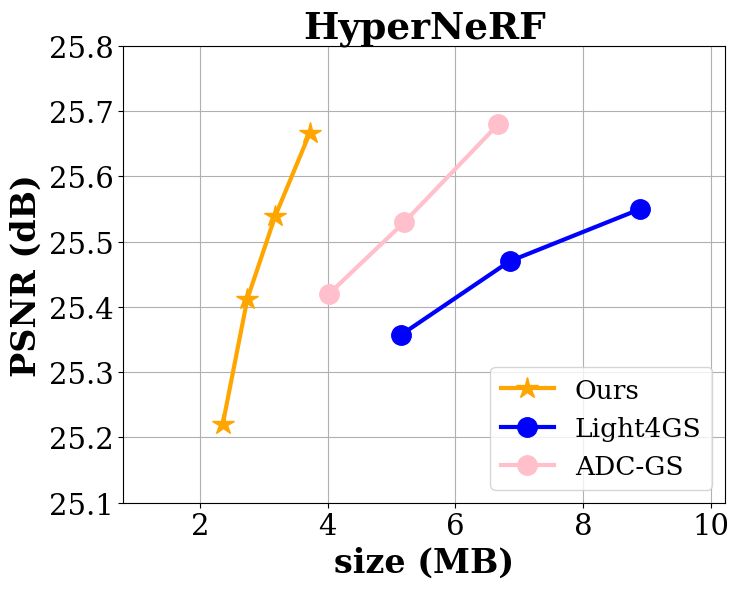}
  \end{subfigure}
  \hfill
  \begin{subfigure}[t]{0.48\columnwidth}
      \centering
      \includegraphics[width=\linewidth]{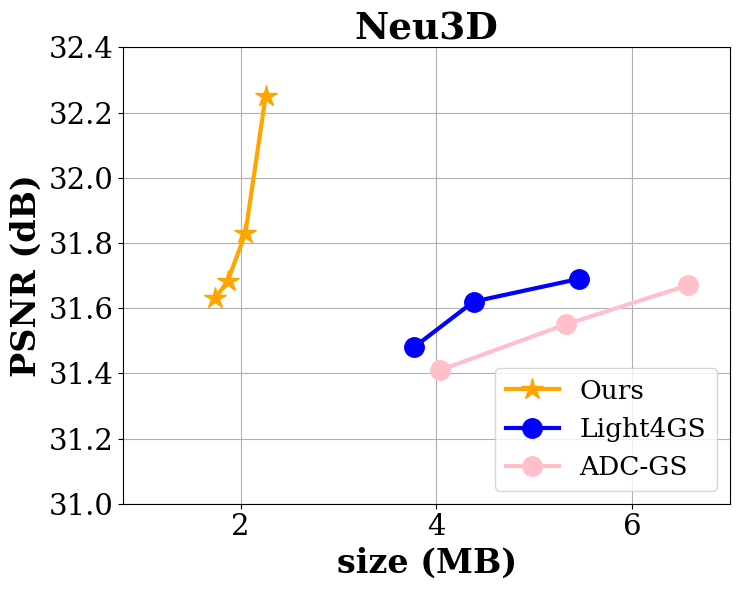}
      % (b)
  \end{subfigure}
  \vspace{-0.5em}
  \caption{Rate-distortion comparison of our TED-4DGS, Light4GS~\cite{liu2025light4gslightweightcompact4d} and ADC-GS~\cite{huang2025adcgsanchordrivendeformablecompressed}.
}
  \label{fig:rd_curve}
  \vspace{-0.5em}
\end{figure}

%% file: sec/exp/quantitive_results.tex
\newcommand{\first}[1]{\colorbox{red!20}{#1}}  % 最高值 - 红色背景
\newcommand{\second}[1]{\colorbox{yellow!20}{#1}} % 第二高值 - 蓝色背景

\begin{table}[t]
\caption{Quantitative results on Neu3D. The best and 2nd best results are highlighted in \first{red} and \second{yellow} cells.}
\label{tab:neu3d-results}
\vspace{-0.75em}
\centering
\resizebox{\columnwidth}{!}{%
\begin{tabular}{l|ccccc}
\hline
\textbf{Model} & \textbf{PSNR$\uparrow$} & \textbf{SSIM$_2$$\uparrow$} & \textbf{LPIPS$\downarrow$} & \textbf{FPS$\uparrow$} & \textbf{Size (MB)$\downarrow$} \\
\hline
4DGS~\cite{yang2024realtimephotorealisticdynamicscene}  & 32.01 & \first{0.986} & 0.055  & \second{114}   & 1000 \\
STG~\cite{li2024spacetimegaussianfeaturesplatting}            & 32.04 & 0.974 & 0.044 & 110   & 175 \\
FreeTimeGS~\cite{wang2025freetimegsfreegaussianprimitives}     & \first{33.19} & --  &  \second{0.036}  & --  &125   \\
\hline 
% 3DGStream      & 31.67 & --    & --                     & 215 & 2340 \\
% DN-4DGS        & 32.02 & 0.984 & 0.043 & 15    & 112 \\
4DGaussians~\cite{wu20244dgaussiansplattingrealtime}     & 31.72 & \second{0.984} & 0.049                  & 34    & 38 \\
E-D3DGS~\cite{bae2024pergaussianembeddingbaseddeformationdeformable}        & 31.20 & 0.974 & \first{0.030} & 42    & 40 \\
\hline 
Light4GS~\cite{liu2025light4gslightweightcompact4d} (low rate)  & 31.48 & -- &  0.064 & 40  & 3.77 \\
Light4GS~\cite{liu2025light4gslightweightcompact4d} (high rate)  & 31.69 & -- & 0.053 & 37 &  5.46  \\
ADC-GS ~\cite{huang2025adcgsanchordrivendeformablecompressed} (low rate)  & 31.41 & 0.972 & 0.066     & \first{126} & 4.04 \\
ADC-GS ~\cite{huang2025adcgsanchordrivendeformablecompressed} (high rate)  & 31.67 & 0.981 & 0.061 & 110 & 6.57 \\
\hline              
\textbf{Ours (low rate)} & 31.63  &0.969 &0.061  &78  &\first{1.73}  \\
\textbf{Ours (high rate)}    &\second{32.25}  &0.972  &0.051  &73  & \second{2.26}   \\
\hline
\end{tabular}
}
\end{table}

\begin{table}[t]
\centering
\caption{Quantitative results on HyperNeRF.}
\label{tab:hypernerf-results}
\vspace{-0.75em}
\resizebox{\columnwidth}{!}{%
\begin{tabular}{l|ccccc}
\hline
\textbf{Model} & \textbf{PSNR$\uparrow$} & \textbf{SSIM$_2$$\uparrow$} & \textbf{LPIPS$\downarrow$} & \textbf{FPS$\uparrow$} & \textbf{Size (MB)$\downarrow$} \\
\hline
% Nerfies        & 22.20 & 0.803 & 0.170 & $<$1  & --  \\
% HyperNeRF      & 22.40 & 0.814 & 0.153 & $<$1  & 15  \\
% TiNeuVox-B     & 24.30 & 0.836 & 0.393                   & 1     & 48  \\
% \hline
% 3DGS           & 19.70 & 0.680 & 0.383                   & 55    & 52  \\
% D3DGS          & 22.40 & 0.612 & 0.275                   & 22    & 129 \\
% DN-4DGS        & 25.59 & 0.861 & --   & 20    & 68  \\
4DGaussians~\cite{wu20244dgaussiansplattingrealtime}     & 25.60 & \first{0.848} & 0.281 & 22    &64  \\
E-D3DGS~\cite{bae2024pergaussianembeddingbaseddeformationdeformable}        & \first{25.74} &0.816 & \first{0.231} & 26    & 47  \\
\hline
Light4GS~\cite{liu2025light4gslightweightcompact4d} (low rate)  & 25.35  &--& --  & 28  & 5.15  \\
Light4GS~\cite{liu2025light4gslightweightcompact4d} (high rate)  &  25.55 &--& --  &  27 &  8.87  \\
ADC-GS ~\cite{huang2025adcgsanchordrivendeformablecompressed} (low rate)  & 25.42 & 0.777 & 0.315 & \first{135} & 4.02 \\
ADC-GS ~\cite{huang2025adcgsanchordrivendeformablecompressed} (high rate)  & \second{25.68} & \second{0.825} & \second{0.252} & 101 & 6.67 \\
\hline 
\textbf{Ours (low rate)}   &25.22    &0.800&0.311&\second{111}& \first{2.36}  \\
\textbf{Ours (high rate)}   &25.67    &0.808 &0.287 &96 & \second{3.72}  \\
\hline
\end{tabular}
}
\vspace{-1em}
\end{table}

%% file: sec/exp/quality_results_2scene.tex
\begin{figure}[t]
  \setlength{\abovecaptionskip}{0pt}
  \setlength{\belowcaptionskip}{0pt}
  \centering
  % 上面那張
  \begin{subfigure}{\linewidth}
    \centering
    \includegraphics[width=\linewidth]{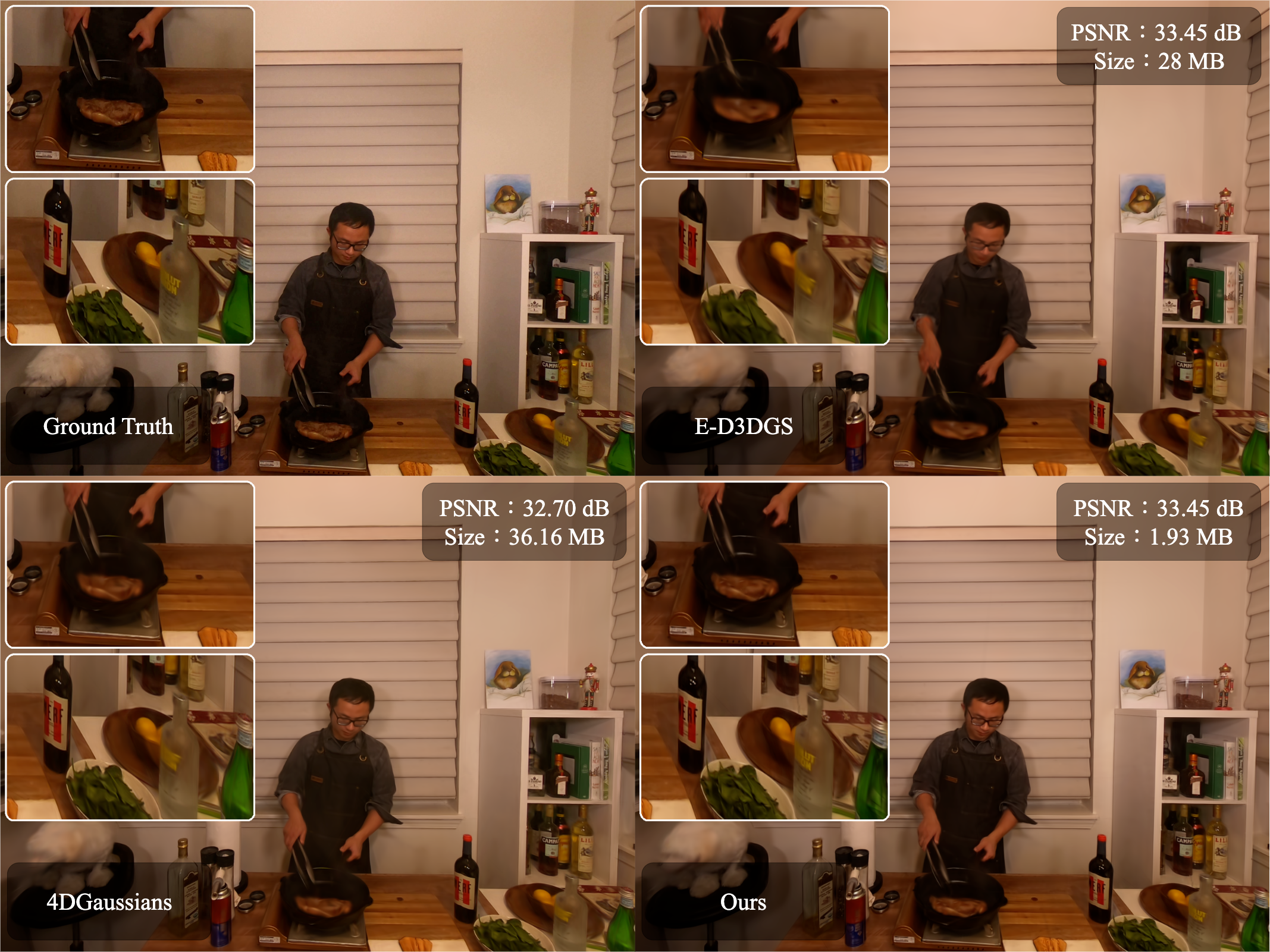}
    % \vspace{-1em}
    \caption{\textit{sear steak} (Neu3D)} % (a)
    \vspace{0.25em}
    \label{fig:hypernerf_render_a}
  \end{subfigure}
%   \vspace{1pt} % 控制上下圖間距

  % 下面那張
  \begin{subfigure}{\linewidth}
    \centering
    \includegraphics[width=\linewidth]{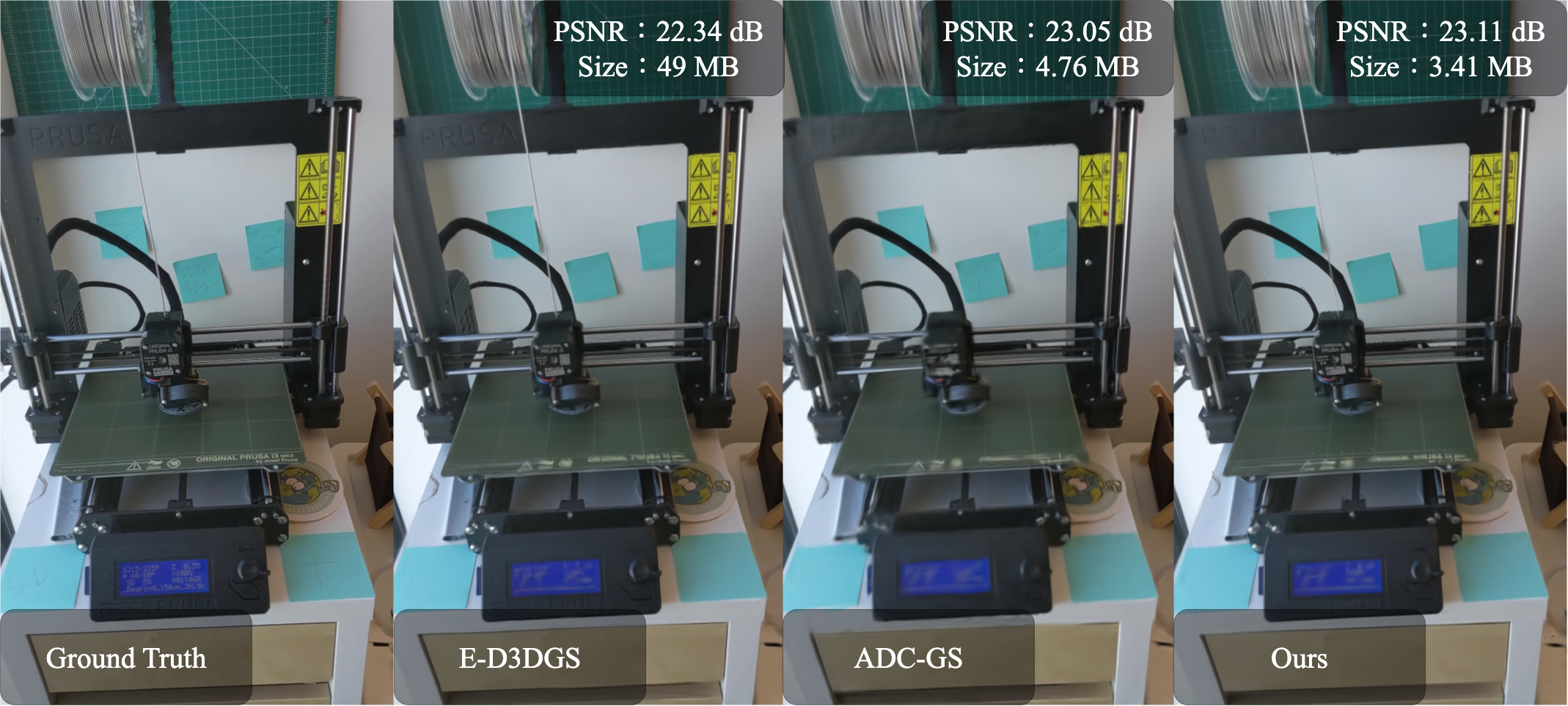}
    % \vspace{-1em}  % 添加這一行
    \caption{\textit{3D printer} (HyperNeRF)} % (b)
    \label{fig:hypernerf_render_b}
  \end{subfigure}
  \vspace{-0.75em}

  \caption{Subjective quality comparisons.}
  % on (a) \textit{sear steak} in Neu3D dataset. (b) \textit{3D printer} in HyperNeRF dataset.}
  \label{fig:render}
  \vspace{-0.5em}
\end{figure}

%% file: sec/exp/5_1_ablation.tex
\subsection{Ablation Experiments}
\label{sec:exp/ablation}

\input{sec/exp/ablation_rd}

\input{sec/exp/pointcloud_compare}
\input{sec/exp/ablation_table.tex}
 \input{sec/exp/temporal_duration_table}

We conduct ablation experiments on the Neu3D~\cite{li2022neural3dvideosynthesis} dataset to validate (1) the query mechanism of the deformation bank, (2) the temporal activation, (3) the progressive training strategy, and (4) the INR-based hyperprior. In addition, we include two baselines: (5) a pure ScaffoldGS compression without temporal modeling and (6) a naive INR-based deformation. Table~\ref{tab:ablation-results} presents RD results at the highest rate point ($\lambda_{rate}=0.002$) while Fig.~\ref{fig:ablation_rd} presents RD-optimized results in the form of RD curves.
%\paragraph*{\textbf{Efficacy of Deformation Components.}}
%Table~\ref{tab:ablation-results} presents our ablation studies about the design of deformation field. 
%We evaluated our deformation components at 60{,}000 iterations (Sections~\ref{sec:method/deformation} and ~\ref{sec:method/3d_to_4d}) prior to applying compression. This allows us to isolate and analyze the effects of deformation design on rendering quality and efficiency, independent of compression. The corresponding rate-distortion curves are also provided in Fig.~{\ref{fig:rd_curve}}.
\vspace{-0.5em}
\paragraph*{\textbf{Deformation Query.}} When querying the global deformation bank $\boldsymbol{Z}$ (Fig. ~\ref{fig:overview} (e)), there are two query mechanisms: multiplication ($\boldsymbol{w}\cdot\boldsymbol{z}^{(t)}$) or concatenation ($\text{concat}(\boldsymbol{\phi}, \boldsymbol{z^{(t)}})$). Our approach adopts the multiplication strategy, which can be interpreted as retrieving relevant deformation from the global deformation vector $\boldsymbol{z}^{(t)}$. From Fig.~\ref{fig:rd_deform}, this approach outperforms the concatenation design ((i) w/ concatenation) in terms of RD performance. 
%Our approach Compared to concatenating anchor and temporal embeddings, our deformation query design, which uses a per-anchor temporal feature to query a global deformation bank, achieves a 0.36\,dB gain under similar storage cost (Table~\ref{tab:ablation-results}). This approach enables anchor-specific motion modeling and better aligns with 4D decomposition principles, as the query structure resembles reconstructing a 4D field via an outer product between 3D spatial and 1D temporal components. 

 % In Table \ref{tab:ablation-results}, (a) we investigate the impact of temporal mask. The temporal mask design can identify between static and dynamic anchors, by doing that, we can skip the temporal feature so as to decrease the total size by 1.4MB. (b) By replacing the 4D formulation with 3D anchor representation, though the PSNR increases by 0.1 dB, the number of anchors also increases significantly by 40\%, leading to the total size increase form 15.18 MB to 20.12 MB.  Furthermore, by cancelling the progressive training, the number of anchors can decrease by 8k, but the PSNR also drops by 0.4 dB, showing that the 4D formulation is established on a well deformation modeling, or it will only pruning more anchors.

 % \paragraph*{\textbf{(b) Temporal mask.}} Removing the temporal mask means that all anchors are regarded as dynamic anchors, leading to a 1.4,MB increase in storage, as all anchors must retain temporal features. With the mask, static anchors can skip temporal encoding, yielding more compact representations with minimal quality loss.

 \vspace{-0.5em}
 \paragraph*{\textbf{Temporal Activation.}} Fig.~\ref{fig:rd_deform} shows that disabling temporal activation reduces our dynamic scene representation with deformation (Section ~\ref{sec:method/3d_to_4d}). This consistently increases the bitrate across all rate points, while the rendering quality exhibits a slight degradation. Recall that our temporal activation approach facilitates time-aware opacity pruning, which discards anchors with limited relevance across time. Fig. ~\ref{fig:3d_4d_pc} further confirms that in the absence of our temporal activation mechanism, a non-trivial yet improper solution may emerge, relocating non-contributing dynamic Gaussian primitives (red dots in Fig.~\ref{fig:3d_4d_pc}) outside the viewing frustum.

We also analyze the active time span of Gaussians ($\Delta \tau = a_f - a_s$). Table~\ref{tab:ablation_temp_dur} and Fig.~\ref{fig:ablation_dur_map_3scenes} show that slow-motion scenes have over 95\% of Gaussians active throughout the sequence, while higher motion complexity produces more short-duration Gaussians with a relatively uniform distribution. This demonstrates that temporal activation adaptively aligns anchor lifetimes with scene dynamics. Further results are provided in Supplementary Sec.~E.

 % and Fig.~\ref{fig:ablation_dur_map_3scenes}

 %This leads to noisier and less stable deformation, as the model is forced to push irrelevant anchors outside the viewing frustum instead of suppressing them temporally. 
 
 % By contrast, the 4D formulation enables smoother and more consistent deformation by explicitly modeling temporal visibility, and it also facilitates temporally-aware opacity pruning to discard anchors with limited temporal relevance, thereby reducing redundancy and enhancing representation compactness. 
 % \vspace{-0.5em}
 \paragraph*{\textbf{Progressive Learning.}} Fig.~\ref{fig:rd_deform} further shows that our progressive training strategy (Section ~\ref{sec:method/3d_to_4d}) leads to superior rate–distortion performance. Across all rate points, we observe a consistent reduction in bitrate, but this comes with a noticeable drop in PSNR. The training process tends to prune anchors more aggressively when their deformation is not well learned in the early phase.
This validates the effectiveness of using static 3D anchors to stabilize deformation learning prior to incorporating temporal activation modeling.
\input{sec/exp/temp_dur_map_3scenes}
\vspace{-0.5em}
\paragraph*{\textbf{INR-based Hyperprior.}}
Fig.~\ref{fig:rd_compress} presents the rate-distortion comparison with a factorized prior~\cite{ballé2017endtoendoptimizedimagecompression} and a grid-based vector-matrix hyperprior~\cite{zhan2025cat3dgspronewbenchmark}. The former treats attribute components as independent and identically distributed random variables. Our approach achieves a 20.0\% BD-rate saving compared to FM and also provide superior RD performance than VM hyperprior.
\vspace{-0.5em}

\paragraph*{\textbf{INR-based Deformation.}} 
In Fig.~\ref{fig:rd_deform}, we also include INR-based deformation as a baseline method, which extends Scaffold-GS ~\cite{lu2023scaffoldgsstructured3dgaussians} with a INR-based deformation network from D3DGS~\cite{yang2023deformable3dgaussianshighfidelity} to model anchor deformation $(\Delta \boldsymbol{x}, \Delta \boldsymbol{f})$ as a function of $(x, y, z, t)$. The baseline performs notably worse than TED-4DGS in rate-distortion performance, highlighting the robustness of our deformation networks.

% achieves a 20.0\% BD-rate saving compared to FM and also provide superior RD performance than one of the SOTA grid-based vector-matrix hyperprior ~\cite{zhan2025cat3dgspronewbenchmark}.  This improvement demonstrates the effectiveness of our hybrid compression strategy.

%% file: sec/exp/ablation_rd.tex
% \begin{figure}[t]
%     \centering
%     \resizebox{\linewidth}{!}{
%     \includegraphics[width=0.5\textwidth]{Figures/exp/image.png}
%     }
%     \caption{%
%     Rate-distortion ablations. \emph{Left:} deformation field variants (ours versus concatenation, w/o temporal activation, w/o progressive training). \emph{Right:} compression variants (ours vs.\ factorized model and grid-based vector-matrix(VM) hyperprior from ~\cite{zhan2025cat3dgspronewbenchmark}).%
%     }
%     \label{fig:ablation_rd}
% \end{figure}

% % \usepackage{subcaption}  % 放在導言區，graphicx 之後

\begin{figure}[t]
  \centering
  \begin{subfigure}[t]{0.48\columnwidth}
      \centering
      \includegraphics[width=\linewidth]{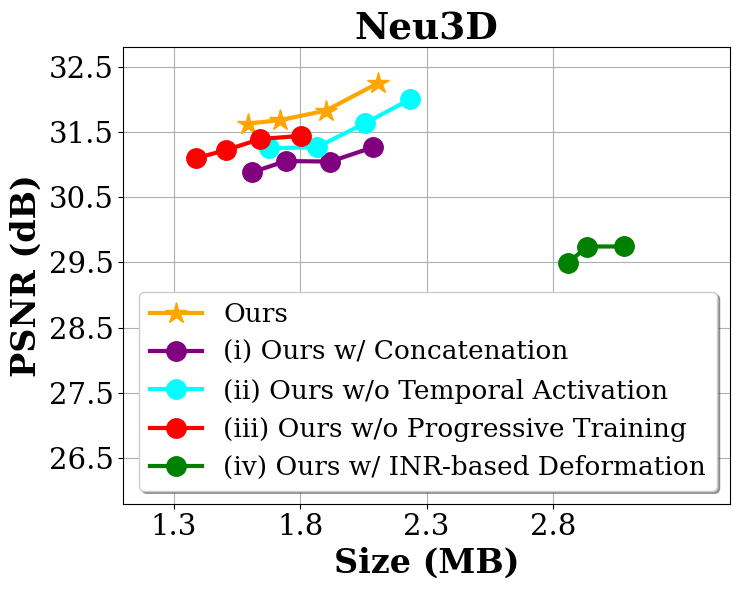}
      \caption{Deformation Field Variants}  % (a)
      \label{fig:rd_deform}
  \end{subfigure}
  \hfill
  \begin{subfigure}[t]{0.48\columnwidth}
      \centering
      \includegraphics[width=\linewidth]{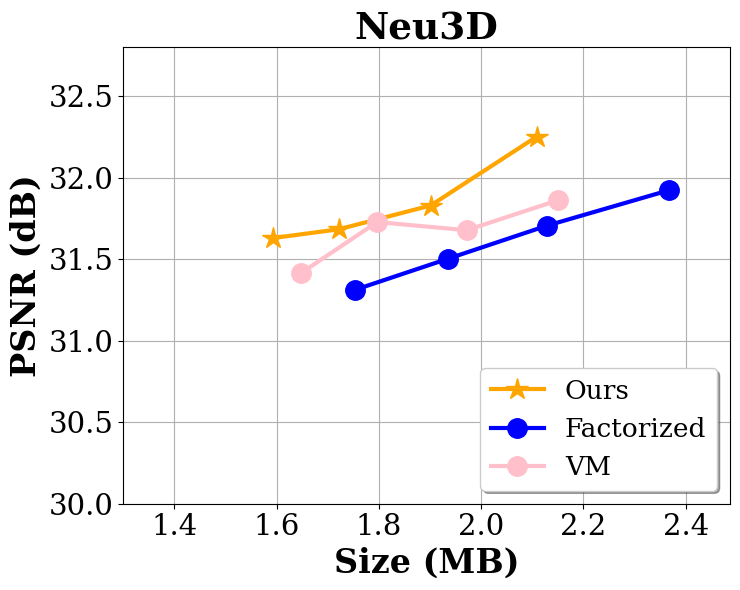}
      \caption{Compression Variants} 
      \label{fig:rd_compress}
      % (b)
  \end{subfigure}
  \vspace{-0.5em}
  \caption{Rate-distortion comparisons on (a) deformation field variants and (b) compression variants.
}
  \label{fig:ablation_rd}
  \vspace{-0.5em}
\end{figure}

%% file: sec/exp/pointcloud_compare.tex
\begin{figure}[t]
% \vspace{-1em}
  \centering
  \begin{subfigure}[t]{0.42\columnwidth}
      \centering
      \includegraphics[width=\linewidth]{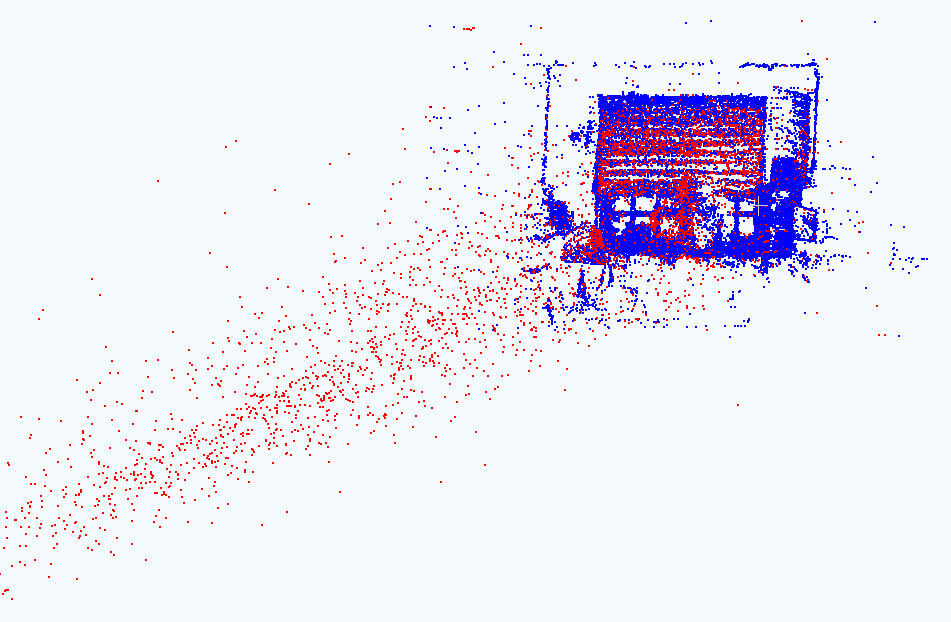}
      \caption{w/o Temporal Activation}  % (a)
  \end{subfigure}
  \hfill
  \begin{subfigure}[t]{0.42\columnwidth}
      \centering
      \includegraphics[width=\linewidth]{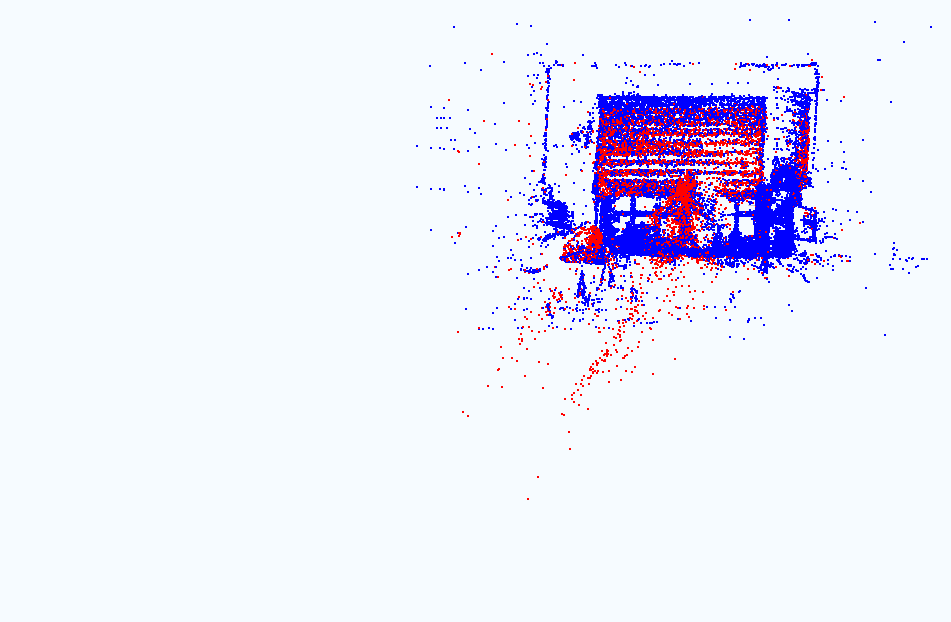}
      \caption{w/ Temporal Activation}                       % (b)
  \end{subfigure}
  \vspace{-0.5em}
  \caption{Comparison of deformed point clouds in the \textit{cook spinach} scene (Neu3D).
  \vspace{-0.5em}
}
  \label{fig:3d_4d_pc}
\end{figure}

%% file: sec/exp/ablation_table.tex
\begin{table}[]
\centering
\caption{Ablation study of deformation field variants on Neu3D at highest rate point.}
\vspace{-0.75em}
\label{tab:ablation-results}
\resizebox{\columnwidth}{!}{%
    \begin{tabular}{l|cc}
    \hline
    \textbf{Method} & \textbf{PSNR $\uparrow$} & \textbf{Size (MB)$\downarrow$} \\
    \hline
    Ours & {32.25} & {2.11} \\
    \hline
    (i) Ours w/ Concatenation& {31.27} & {2.10} \\
    \hline
    (ii) Ours w/o Temporal Activation& {32.00} & {2.23}\\
    \hline
    (iii) Ours w/o Progressive Training& {31.44} & {1.80} \\
    \hline
     (iv) {Ours w/ INR-based Deformation} & {29.72} & {5.09} \\
     \hline
     % (v) Scaffold-GS w/o deformation&28.16 & 17.33\\
     % \hline
    \end{tabular}
}
\end{table}

% \begin{table}[]
% \centering
% \caption{Ablation study of deformation field variants on Neu3D.}
% \vspace{-1em}
% \label{tab:ablation-results}
% \resizebox{\columnwidth}{!}{%
%     \begin{tabular}{l|ccc}
%     \hline
%     \textbf{Method} & \textbf{PSNR $\uparrow$} & \textbf{Size (MB)$\downarrow$} & \textbf{Anchors \#  (K)$\downarrow$}\\
%     \hline
%     Ours & 31.86 & 15.18 & 40\\
%     \hline
%     (i) Ours w/ Concatenation& 31.50 & 16.71 & 44\\
%     \hline
%     (ii) Ours w/o Temporal Activation& 32.03 & 20.12 & 55\\
%     \hline
%     (iii) Ours w/o Progressive Training& 31.48 & 12.49 & 32\\
%     \hline
%      (iv) Scaffold-GS + INR-based Deformation&29.77 &23.51 & 60\\
%      \hline
%      (v) Scaffold-GS w/o deformation&28.16 & 17.33&49\\
%      \hline
%     \end{tabular}
% }
% \end{table}

%% file: sec/exp/temporal_duration_table.tex
\begin{table}[t]
\centering
\caption{{Temporal duration versus motion complexity.}}
\vspace{-0.75em}
\label{tab:ablation_temp_dur}
\resizebox{\columnwidth}{!}{%
    \begin{tabular}{llccc}\toprule
    
     Scene&Motion & \multicolumn{3}{c}{Temporal Duration ($\Delta \tau=a_f-a_s$)}\\
    
     &Complexity& $\Delta \tau \le 0.2$& $0.2<\Delta \tau < 0.8$& $0.8 \le \Delta \tau$\\\midrule
    
     Flame Steak&Slow& 3\%& 0\%& 97\%\\
    
     Banana&Medium& 9\%& 6\%& 85\%\\
    
     Broom&Fast& 18\%& 35\%& 47\%\\ \bottomrule
    \end{tabular}
}
\end{table}

%% file: sec/exp/temp_dur_map_3scenes.tex
\begin{figure}[t]
  \centering
  \includegraphics[width=0.43\textwidth]{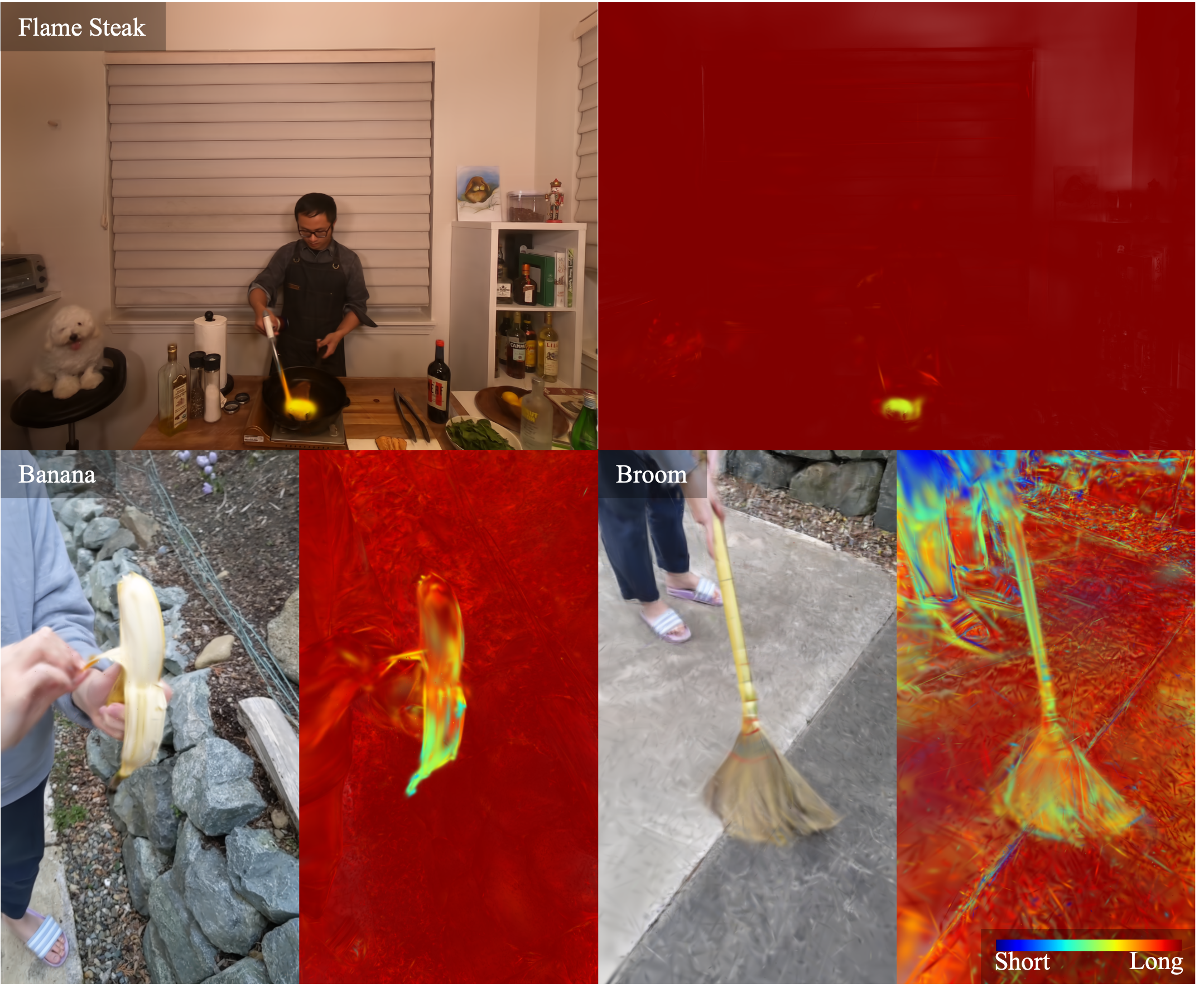}
  \vspace{-0.75em}
  \captionof{figure}{{Comparison of rendered images and rendered temporal duration maps with different motion dynamics.}}
  \vspace{-1em}
  \label{fig:ablation_dur_map_3scenes}
\end{figure}

%% file: sec/6_conclusion.tex
\section{Conclusion}
This work presents a novel rate-distortion-optimized compression framework for dynamic 3DGS. To efficiently represent anchor-wise temporal deformation, it features a compact embedding-based deformation network that leverages per-anchor temporal embeddings and a shared global deformation bank. Furthermore, a progressive 4D anchor learning strategy is introduced to ensure smooth deformation and explicit temporal visibility control via learnable temporal activation parameters. Finally, to achieve effective attribute coding, we combine an INR-based hyperprior and a channel-wise autoregressive model to predict the distribution for entropy coding. These components collectively enable TED-4DGS to achieve state-of-the-art rate-distortion performance across several real-world dynamic scene datasets.

\section{Acknowledgement}
This work is supported by MediaTek Advanced Research Center and National Science and Technology Council (NSTC), Taiwan, under Grants 113-2634-F-A49-007-, 112-2221-E-A49-092-MY3, and 114-2221-E-A49-035-MY3. We thank to National Center for High-performance Computing (NCHC) for providing computational and storage resources.

%% file: main.bbl
\begin{thebibliography}{46}
\providecommand{\natexlab}[1]{#1}
\providecommand{\url}[1]{\texttt{#1}}
\expandafter\ifx\csname urlstyle\endcsname\relax
  \providecommand{\doi}[1]{doi: #1}\else
  \providecommand{\doi}{doi: \begingroup \urlstyle{rm}\Url}\fi

\bibitem[Ali et~al.(2024)Ali, Qamar, Bae, and Tartaglione]{ali2024trimmingfatefficientcompression}
Muhammad~Salman Ali, Maryam Qamar, Sung-Ho Bae, and Enzo Tartaglione.
\newblock Trimming the fat: Efficient compression of 3d gaussian splats through pruning, 2024.

\bibitem[Bae et~al.(2024)Bae, Kim, Yun, Lee, Bang, and Uh]{bae2024pergaussianembeddingbaseddeformationdeformable}
Jeongmin Bae, Seoha Kim, Youngsik Yun, Hahyun Lee, Gun Bang, and Youngjung Uh.
\newblock Per-gaussian embedding-based deformation for deformable 3d gaussian splatting, 2024.

\bibitem[Ballé et~al.(2017)Ballé, Laparra, and Simoncelli]{ballé2017endtoendoptimizedimagecompression}
Johannes Ballé, Valero Laparra, and Eero~P. Simoncelli.
\newblock End-to-end optimized image compression, 2017.

\bibitem[Ballé et~al.(2018)Ballé, Minnen, Singh, Hwang, and Johnston]{ballé2018variationalimagecompressionscale}
Johannes Ballé, David Minnen, Saurabh Singh, Sung~Jin Hwang, and Nick Johnston.
\newblock Variational image compression with a scale hyperprior, 2018.

\bibitem[Cao and Johnson(2023)]{cao2023hexplanefastrepresentationdynamic}
Ang Cao and Justin Johnson.
\newblock Hexplane: A fast representation for dynamic scenes, 2023.

\bibitem[Chen et~al.(2024)Chen, Wu, Lin, Harandi, and Cai]{chen2024hachashgridassistedcontext}
Yihang Chen, Qianyi Wu, Weiyao Lin, Mehrtash Harandi, and Jianfei Cai.
\newblock Hac: Hash-grid assisted context for 3d gaussian splatting compression, 2024.

\bibitem[Chen et~al.(2025)Chen, Wu, Li, Lin, Harandi, and Cai]{chen2025fastfeedforward3dgaussian}
Yihang Chen, Qianyi Wu, Mengyao Li, Weiyao Lin, Mehrtash Harandi, and Jianfei Cai.
\newblock Fast feedforward 3d gaussian splatting compression, 2025.

\bibitem[Cho et~al.(2024)Cho, Cho, Kim, Bae, Uh, and Kim]{cho20244dscaffoldgaussiansplatting}
Woong~Oh Cho, In Cho, Seoha Kim, Jeongmin Bae, Youngjung Uh, and Seon~Joo Kim.
\newblock 4d scaffold gaussian splatting for memory efficient dynamic scene reconstruction, 2024.

\bibitem[Duan et~al.(2024)Duan, Wei, Dai, He, Chen, and Chen]{duan20244drotorgaussiansplattingefficient}
Yuanxing Duan, Fangyin Wei, Qiyu Dai, Yuhang He, Wenzheng Chen, and Baoquan Chen.
\newblock 4d-rotor gaussian splatting: Towards efficient novel view synthesis for dynamic scenes, 2024.

\bibitem[Duisterhof et~al.(2024)Duisterhof, Mandi, Yao, Liu, Seidenschwarz, Shou, Ramanan, Song, Birchfield, Wen, and Ichnowski]{duisterhof2024deformgssceneflowhighly}
Bardienus~P. Duisterhof, Zhao Mandi, Yunchao Yao, Jia-Wei Liu, Jenny Seidenschwarz, Mike~Zheng Shou, Deva Ramanan, Shuran Song, Stan Birchfield, Bowen Wen, and Jeffrey Ichnowski.
\newblock Deformgs: Scene flow in highly deformable scenes for deformable object manipulation, 2024.

\bibitem[Fan et~al.(2025)Fan, Chang, Liu, Lee, Huang, Tseng, and Liu]{fan2025spectromotiondynamic3dreconstruction}
Cheng-De Fan, Chen-Wei Chang, Yi-Ruei Liu, Jie-Ying Lee, Jiun-Long Huang, Yu-Chee Tseng, and Yu-Lun Liu.
\newblock Spectromotion: Dynamic 3d reconstruction of specular scenes, 2025.

\bibitem[Fan et~al.(2024)Fan, Wang, Wen, Zhu, Xu, and Wang]{fan2024lightgaussianunbounded3dgaussian}
Zhiwen Fan, Kevin Wang, Kairun Wen, Zehao Zhu, Dejia Xu, and Zhangyang Wang.
\newblock Lightgaussian: Unbounded 3d gaussian compression with 15x reduction and 200+ fps, 2024.

\bibitem[Fischer et~al.(2024)Fischer, Kulhanek, Bulò, Porzi, Pollefeys, and Kontschieder]{fischer2024dynamic3dgaussianfields}
Tobias Fischer, Jonas Kulhanek, Samuel~Rota Bulò, Lorenzo Porzi, Marc Pollefeys, and Peter Kontschieder.
\newblock Dynamic 3d gaussian fields for urban areas, 2024.

\bibitem[He et~al.(2022)He, Yang, Peng, Ma, Qin, and Wang]{he2022elicefficientlearnedimage}
Dailan He, Ziming Yang, Weikun Peng, Rui Ma, Hongwei Qin, and Yan Wang.
\newblock Elic: Efficient learned image compression with unevenly grouped space-channel contextual adaptive coding, 2022.

\bibitem[Huang et~al.(2025)Huang, Yang, Liu, Xu, and Li]{huang2025adcgsanchordrivendeformablecompressed}
He Huang, Qi Yang, Mufan Liu, Yiling Xu, and Zhu Li.
\newblock Adc-gs: Anchor-driven deformable and compressed gaussian splatting for dynamic scene reconstruction, 2025.

\bibitem[Huang et~al.(2024)Huang, Sun, Yang, Lyu, Cao, and Qi]{huang2024scgssparsecontrolledgaussiansplatting}
Yi-Hua Huang, Yang-Tian Sun, Ziyi Yang, Xiaoyang Lyu, Yan-Pei Cao, and Xiaojuan Qi.
\newblock Sc-gs: Sparse-controlled gaussian splatting for editable dynamic scenes, 2024.

\bibitem[Jiawei et~al.(2024)Jiawei, Zexin, Jian, and Jin]{xu2024grid4d}
Xu Jiawei, Fan Zexin, Yang Jian, and Xie Jin.
\newblock {Grid4D}: {4D} decomposed hash encoding for high-fidelity dynamic gaussian splatting.
\newblock \emph{The Thirty-eighth Annual Conference on Neural Information Processing Systems}, 2024.

\bibitem[Kong et~al.(2025)Kong, Yang, and Wang]{kong2025efficientgaussiansplattingmonocular}
Hanyang Kong, Xingyi Yang, and Xinchao Wang.
\newblock Efficient gaussian splatting for monocular dynamic scene rendering via sparse time-variant attribute modeling, 2025.

\bibitem[Kratimenos et~al.(2024)Kratimenos, Lei, and Daniilidis]{kratimenos2024dynmfneuralmotionfactorization}
Agelos Kratimenos, Jiahui Lei, and Kostas Daniilidis.
\newblock Dynmf: Neural motion factorization for real-time dynamic view synthesis with 3d gaussian splatting, 2024.

\bibitem[Kwak et~al.(2025)Kwak, Kim, Jeong, Cheong, Oh, and Kim]{kwak2025modecgsglobaltolocalmotiondecomposition}
Sangwoon Kwak, Joonsoo Kim, Jun~Young Jeong, Won-Sik Cheong, Jihyong Oh, and Munchurl Kim.
\newblock Modec-gs: Global-to-local motion decomposition and temporal interval adjustment for compact dynamic 3d gaussian splatting, 2025.

\bibitem[Labe et~al.(2024)Labe, Issachar, Lang, and Benaim]{labe2024dgddynamic3dgaussians}
Isaac Labe, Noam Issachar, Itai Lang, and Sagie Benaim.
\newblock Dgd: Dynamic 3d gaussians distillation, 2024.

\bibitem[Lee et~al.(2024{\natexlab{a}})Lee, Won, Jung, Bae, and Jeon]{lee2024fullyexplicitdynamicgaussian}
Junoh Lee, Chang-Yeon Won, Hyunjun Jung, Inhwan Bae, and Hae-Gon Jeon.
\newblock Fully explicit dynamic gaussian splatting, 2024{\natexlab{a}}.

\bibitem[Lee et~al.(2024{\natexlab{b}})Lee, Rho, Sun, Ko, and Park]{lee2024compact3dgaussianrepresentation}
Joo~Chan Lee, Daniel Rho, Xiangyu Sun, Jong~Hwan Ko, and Eunbyung Park.
\newblock Compact 3d gaussian representation for radiance field, 2024{\natexlab{b}}.

\bibitem[Li et~al.(2022)Li, Slavcheva, Zollhoefer, Green, Lassner, Kim, Schmidt, Lovegrove, Goesele, Newcombe, and Lv]{li2022neural3dvideosynthesis}
Tianye Li, Mira Slavcheva, Michael Zollhoefer, Simon Green, Christoph Lassner, Changil Kim, Tanner Schmidt, Steven Lovegrove, Michael Goesele, Richard Newcombe, and Zhaoyang Lv.
\newblock Neural 3d video synthesis from multi-view video, 2022.

\bibitem[Li et~al.(2024)Li, Chen, Li, and Xu]{li2024spacetimegaussianfeaturesplatting}
Zhan Li, Zhang Chen, Zhong Li, and Yi Xu.
\newblock Spacetime gaussian feature splatting for real-time dynamic view synthesis, 2024.

\bibitem[Liu et~al.(2025{\natexlab{a}})Liu, Chen, Jiang, Wang, and Xu]{liu2025hemgshybridentropymodel}
Lei Liu, Zhenghao Chen, Wei Jiang, Wei Wang, and Dong Xu.
\newblock Hemgs: A hybrid entropy model for 3d gaussian splatting data compression, 2025{\natexlab{a}}.

\bibitem[Liu et~al.(2025{\natexlab{b}})Liu, Yang, Huang, Huang, Yuan, Li, and Xu]{liu2025light4gslightweightcompact4d}
Mufan Liu, Qi Yang, He Huang, Wenjie Huang, Zhenlong Yuan, Zhu Li, and Yiling Xu.
\newblock Light4gs: Lightweight compact 4d gaussian splatting generation via context model, 2025{\natexlab{b}}.

\bibitem[Liu et~al.(2025{\natexlab{c}})Liu, Liu, Wang, Lyv, Wang, Wang, and Hou]{liu2025modgsdynamicgaussiansplatting}
Qingming Liu, Yuan Liu, Jiepeng Wang, Xianqiang Lyv, Peng Wang, Wenping Wang, and Junhui Hou.
\newblock Modgs: Dynamic gaussian splatting from casually-captured monocular videos with depth priors, 2025{\natexlab{c}}.

\bibitem[Liu et~al.(2024)Liu, Wu, Zhang, Wang, Li, and Kwong]{liu2024compgsefficient3dscene}
Xiangrui Liu, Xinju Wu, Pingping Zhang, Shiqi Wang, Zhu Li, and Sam Kwong.
\newblock Compgs: Efficient 3d scene representation via compressed gaussian splatting, 2024.

\bibitem[Lu et~al.(2023)Lu, Yu, Xu, Xiangli, Wang, Lin, and Dai]{lu2023scaffoldgsstructured3dgaussians}
Tao Lu, Mulin Yu, Linning Xu, Yuanbo Xiangli, Limin Wang, Dahua Lin, and Bo Dai.
\newblock Scaffold-gs: Structured 3d gaussians for view-adaptive rendering, 2023.

\bibitem[Matsuki et~al.(2025)Matsuki, Bae, and Davison]{matsuki20254dtamnonrigidtrackingmapping}
Hidenobu Matsuki, Gwangbin Bae, and Andrew~J. Davison.
\newblock 4dtam: Non-rigid tracking and mapping via dynamic surface gaussians, 2025.

\bibitem[Minnen and Singh(2020)]{minnen2020channelwiseautoregressiveentropymodels}
David Minnen and Saurabh Singh.
\newblock Channel-wise autoregressive entropy models for learned image compression, 2020.

\bibitem[Minnen et~al.(2018)Minnen, Ballé, and Toderici]{minnen2018jointautoregressivehierarchicalpriors}
David Minnen, Johannes Ballé, and George Toderici.
\newblock Joint autoregressive and hierarchical priors for learned image compression, 2018.

\bibitem[Müller et~al.(2022)Müller, Evans, Schied, and Keller]{M_ller_2022}
Thomas Müller, Alex Evans, Christoph Schied, and Alexander Keller.
\newblock Instant neural graphics primitives with a multiresolution hash encoding.
\newblock \emph{ACM Transactions on Graphics}, 41\penalty0 (4):\penalty0 1–15, 2022.

\bibitem[Oh et~al.(2025)Oh, Lee, Jeon, and Park]{oh2025hybrid3d4dgaussiansplatting}
Seungjun Oh, Younggeun Lee, Hyejin Jeon, and Eunbyung Park.
\newblock Hybrid 3d-4d gaussian splatting for fast dynamic scene representation, 2025.

\bibitem[Park et~al.(2021)Park, Sinha, Hedman, Barron, Bouaziz, Goldman, Martin-Brualla, and Seitz]{park2021hypernerfhigherdimensionalrepresentationtopologically}
Keunhong Park, Utkarsh Sinha, Peter Hedman, Jonathan~T. Barron, Sofien Bouaziz, Dan~B Goldman, Ricardo Martin-Brualla, and Steven~M. Seitz.
\newblock Hypernerf: A higher-dimensional representation for topologically varying neural radiance fields, 2021.

\bibitem[Shaw et~al.(2024)Shaw, Nazarczuk, Song, Moreau, Catley-Chandar, Dhamo, and Perez-Pellitero]{shaw2024swingsslidingwindowsdynamic}
Richard Shaw, Michal Nazarczuk, Jifei Song, Arthur Moreau, Sibi Catley-Chandar, Helisa Dhamo, and Eduardo Perez-Pellitero.
\newblock Swings: Sliding windows for dynamic 3d gaussian splatting, 2024.

\bibitem[Wang et~al.(2024)Wang, Li, Guo, Yang, Kot, and Wen]{wang2024contextgscompact3dgaussian}
Yufei Wang, Zhihao Li, Lanqing Guo, Wenhan Yang, Alex~C. Kot, and Bihan Wen.
\newblock Contextgs: Compact 3d gaussian splatting with anchor level context model, 2024.

\bibitem[Wang et~al.(2025)Wang, Yang, Xu, Sun, Zhang, Chen, Bao, Peng, and Zhou]{wang2025freetimegsfreegaussianprimitives}
Yifan Wang, Peishan Yang, Zhen Xu, Jiaming Sun, Zhanhua Zhang, Yong Chen, Hujun Bao, Sida Peng, and Xiaowei Zhou.
\newblock Freetimegs: Free gaussian primitives at anytime and anywhere for dynamic scene reconstruction, 2025.

\bibitem[Wu et~al.(2024)Wu, Yi, Fang, Xie, Zhang, Wei, Liu, Tian, and Wang]{wu20244dgaussiansplattingrealtime}
Guanjun Wu, Taoran Yi, Jiemin Fang, Lingxi Xie, Xiaopeng Zhang, Wei Wei, Wenyu Liu, Qi Tian, and Xinggang Wang.
\newblock 4d gaussian splatting for real-time dynamic scene rendering, 2024.

\bibitem[Yang et~al.(2025)Yang, Liu, Cui, Liu, Zhang, Yan, and Wang]{yang2025ntrgaussiannighttimedynamicthermal}
Kun Yang, Yuxiang Liu, Zeyu Cui, Yu Liu, Maojun Zhang, Shen Yan, and Qing Wang.
\newblock Ntr-gaussian: Nighttime dynamic thermal reconstruction with 4d gaussian splatting based on thermodynamics, 2025.

\bibitem[Yang et~al.(2024{\natexlab{a}})Yang, Li, Shen, Gong, Dou, and Jin]{yang2024deform3dgsflexibledeformationfast}
Shuojue Yang, Qian Li, Daiyun Shen, Bingchen Gong, Qi Dou, and Yueming Jin.
\newblock Deform3dgs: Flexible deformation for fast surgical scene reconstruction with gaussian splatting, 2024{\natexlab{a}}.

\bibitem[Yang et~al.(2023)Yang, Gao, Zhou, Jiao, Zhang, and Jin]{yang2023deformable3dgaussianshighfidelity}
Ziyi Yang, Xinyu Gao, Wen Zhou, Shaohui Jiao, Yuqing Zhang, and Xiaogang Jin.
\newblock Deformable 3d gaussians for high-fidelity monocular dynamic scene reconstruction, 2023.

\bibitem[Yang et~al.(2024{\natexlab{b}})Yang, Yang, Pan, and Zhang]{yang2024realtimephotorealisticdynamicscene}
Zeyu Yang, Hongye Yang, Zijie Pan, and Li Zhang.
\newblock Real-time photorealistic dynamic scene representation and rendering with 4d gaussian splatting, 2024{\natexlab{b}}.

\bibitem[Zhan et~al.(2025{\natexlab{a}})Zhan, bi~Yang, Ho, Chiang, and Peng]{zhan2025cat3dgspronewbenchmark}
Yu-Ting Zhan, He bi Yang, Cheng-Yuan Ho, Jui-Chiu Chiang, and Wen-Hsiao Peng.
\newblock Cat-3dgs pro: A new benchmark for efficient 3dgs compression, 2025{\natexlab{a}}.

\bibitem[Zhan et~al.(2025{\natexlab{b}})Zhan, Ho, Yang, Chen, Chiang, Liu, and Peng]{zhan2025cat3dgscontextadaptivetriplaneapproach}
Yu-Ting Zhan, Cheng-Yuan Ho, Hebi Yang, Yi-Hsin Chen, Jui~Chiu Chiang, Yu-Lun Liu, and Wen-Hsiao Peng.
\newblock Cat-3dgs: A context-adaptive triplane approach to rate-distortion-optimized 3dgs compression, 2025{\natexlab{b}}.

\end{thebibliography}
